\documentclass[10pt,twocolumn,letterpaper]{article}

\usepackage{cvpr}      

\usepackage{graphicx}
\usepackage{amsmath}
\usepackage{amssymb}
\usepackage{booktabs}
\usepackage{epsfig}

\usepackage{multirow}
\usepackage{verbatim}
\usepackage[pagebackref,breaklinks,colorlinks]{hyperref}

\usepackage[capitalize]{cleveref}
\crefname{section}{Sec.}{Secs.}
\Crefname{section}{Section}{Sections}
\Crefname{table}{Table}{Tables}
\crefname{table}{Tab.}{Tabs.}

\def\cvprPaperID{7855} 
\def\confName{CVPR}
\def\confYear{2022}

\begin{document}

\title{Pyramid Feature Alignment Network for Video Deblurring}

\author{Leitian Tao \\ Wuhan University
\and  
Zhenzhong Chen
\thanks{Corresponding author: Zhenzhong Chen, email: zzchen@ieee.org.}  \\ Wuhan University
}

\maketitle

\begin{abstract}
    Video deblurring remains a challenging task due to various causes of blurring. Traditional methods have considered how to utilize neighboring frames by the single-scale alignment for restoration. However, they typically suffer from misalignment caused by severe blur.  In this work, we aim to better utilize neighboring frames with high efficient feature alignment. We propose a Pyramid Feature Alignment Network (PFAN) for video deblurring. First, the multi-scale feature of blurry frames is extracted with the strategy of Structure-to-Detail Downsampling (SDD) before alignment. This downsampling strategy makes the edges sharper, which is helpful for alignment. Then we align the feature at each scale and reconstruct the image at the corresponding scale. This strategy effectively supervises the alignment at each scale, overcoming the problem of propagated errors from the above scales at the alignment stage. To better handle the challenges of complex and large motions, instead of aligning features at each scale separately, lower-scale motion information is used to guide the higher-scale motion estimation. Accordingly, a Cascade Guided Deformable Alignment (CGDA) is proposed to integrate coarse motion into deformable convolution for finer and more accurate alignment. As demonstrated in extensive experiments, our proposed PFAN achieves superior performance with competitive speed compared to the state-of-the-art methods.
 
\end{abstract}
\vspace{-4mm}

\section{Introduction}\label[section]{intro}
Videos captured by hand-held cameras are susceptible to undesirable blurs, which could be caused by various reasons such as camera shaking or object motion, \textit{etc}. Blur can affect the quality of the video, degrade viewers' watching experiences, as well as degenerate the performance of some high-level video processing tasks, \textit{e.g.,} tracking \cite{mei2008modeling}, video stabilization \cite{matsushita2006full}, and SLAM~\cite{lee2011simultaneous}. Therefore, video deblurring, which aims to restore the frames from a given blurry video clip, has become an active topic that attracts researchers' attention.

\begin{figure}
  \centering

  \includegraphics[width=0.9\linewidth]{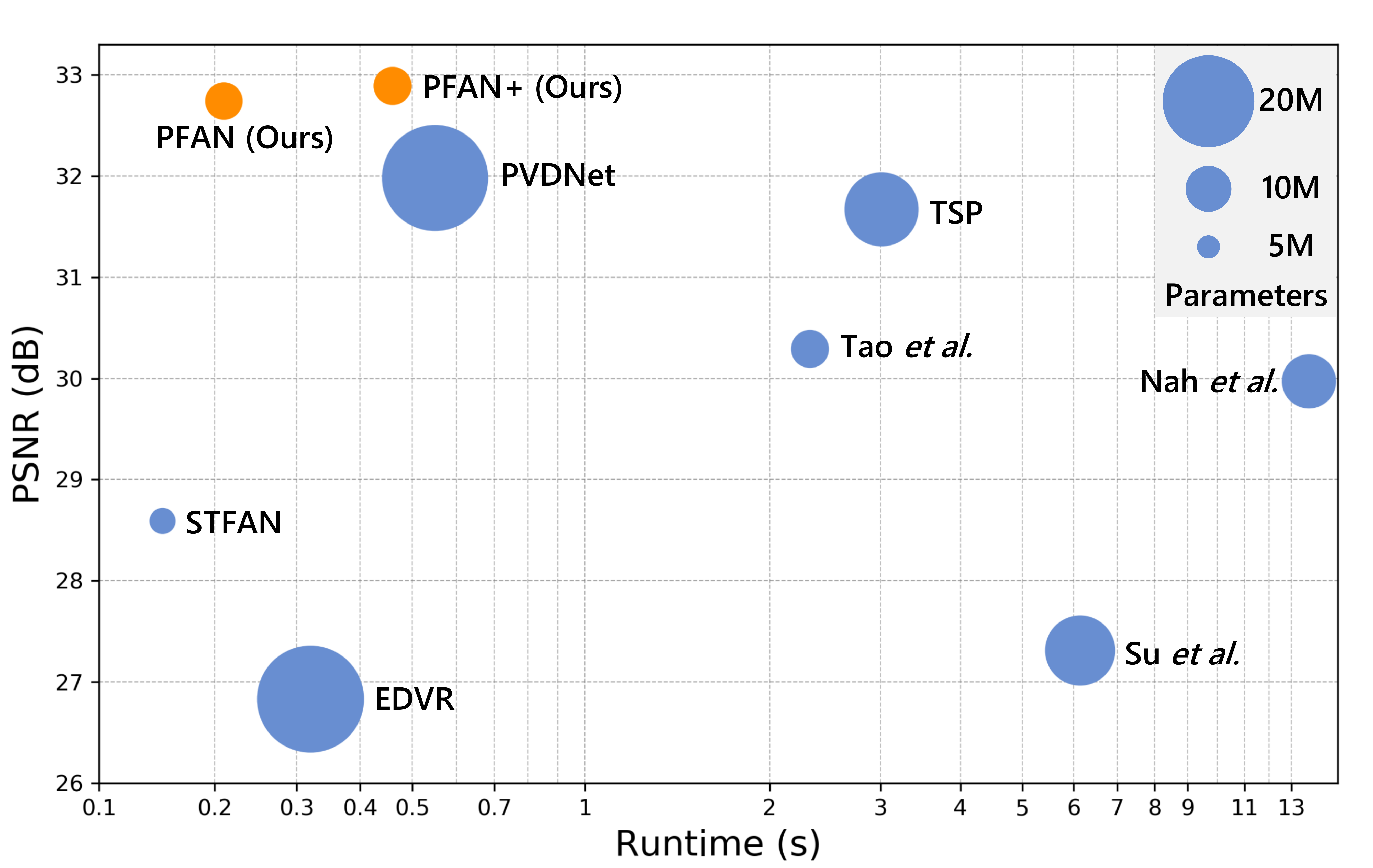}
  \caption{Performance comparisons of video deblurring methods on the GoPro dataset~\cite{nah2017deep}. Our method achieves superior performance in terms of PSNR, while being efficient in terms of running time and model size.
  }
  \label{fig:GoPro}
\vspace{-4mm}
\end{figure}

Tremendous progress has been achieved in single image deblurring because many deblurring convolutional neural networks (CNN) with a coarse-to-fine structure have been introduced and demonstrated to be effective \cite{nah2017deep,tao2018scale,gao2019dynamic,gou2020clearer,cho2021rethinking}. However, video deblurring is a more challenge task as it is difficult to effectively model and utilize the inherent temporal information from the neighboring frames. It is noticed that aligning the blurry images in consecutive video frames has become an effective direction for video deblurring. Previous video deblurring methods mainly focused on aligning frames at a single scale via optical flow ~\cite{pan2020cascaded,su2017deep, kim2018spatio, xue2019video} or deformable convolutions~\cite{zhou2019spatio, wang2019edvr}.  However, the alignment may fail due to occlusion, large motion, and the severity of blurring. Therefore, mainstream methods typically perform alignment in a coarse-to-fine manner, where the misalignment problem can be addressed by aligning frames at multiple scales.  The aligned blurry frames are then proceeded into the subsequent image-deblurring modules, where another round of multi-scale reconstruction is applied for deblurring. 

\begin{figure*}
  \centering
  \begin{subfigure}{0.49\textwidth}
      \includegraphics[width=1\textwidth]{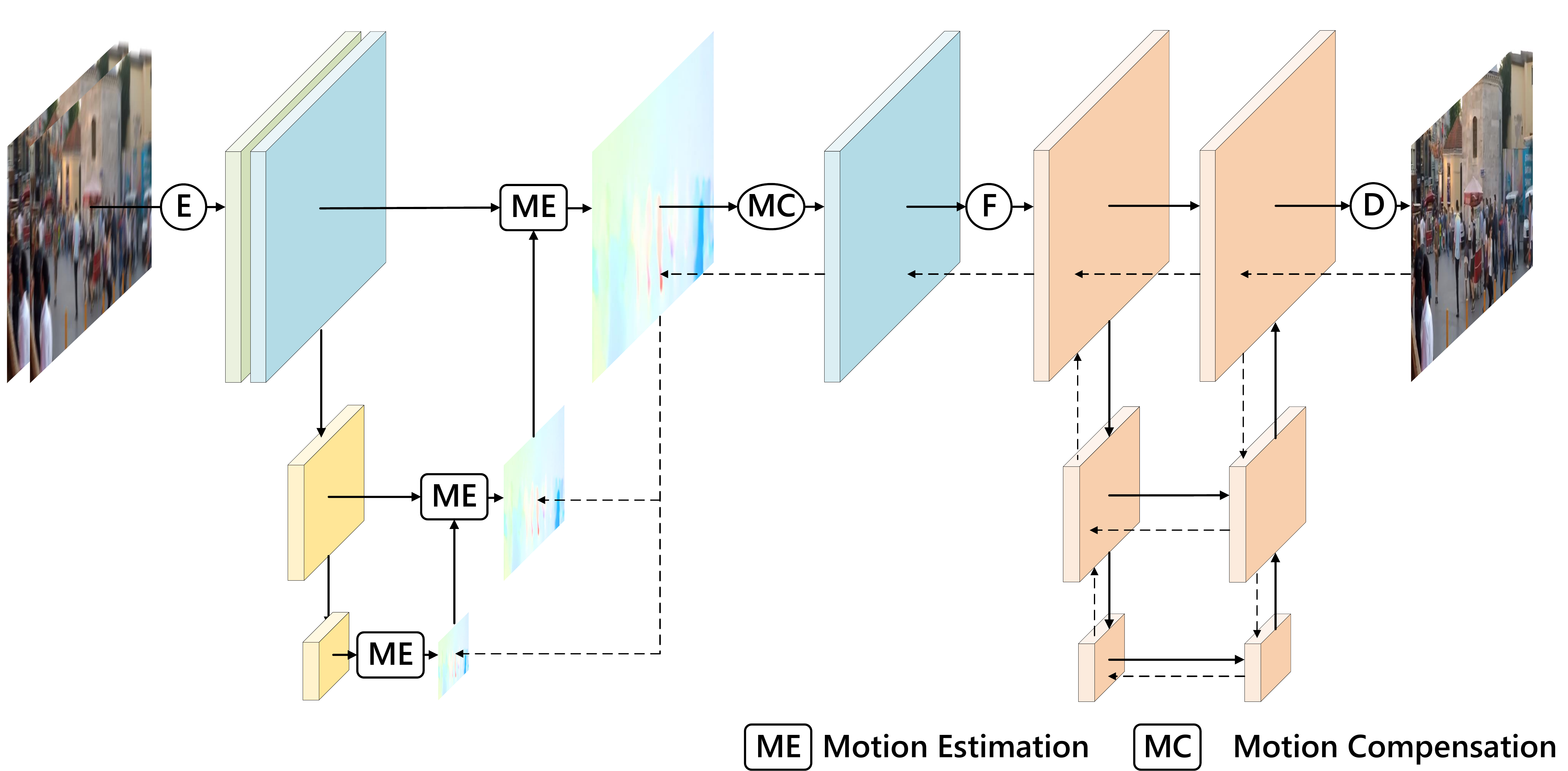}
    \caption{Single Scale Alignment}
    \label{fig:SLA}
  \end{subfigure}
  \hspace{3mm}
  \begin{subfigure}{0.48\textwidth}
    \includegraphics[width=1\textwidth]{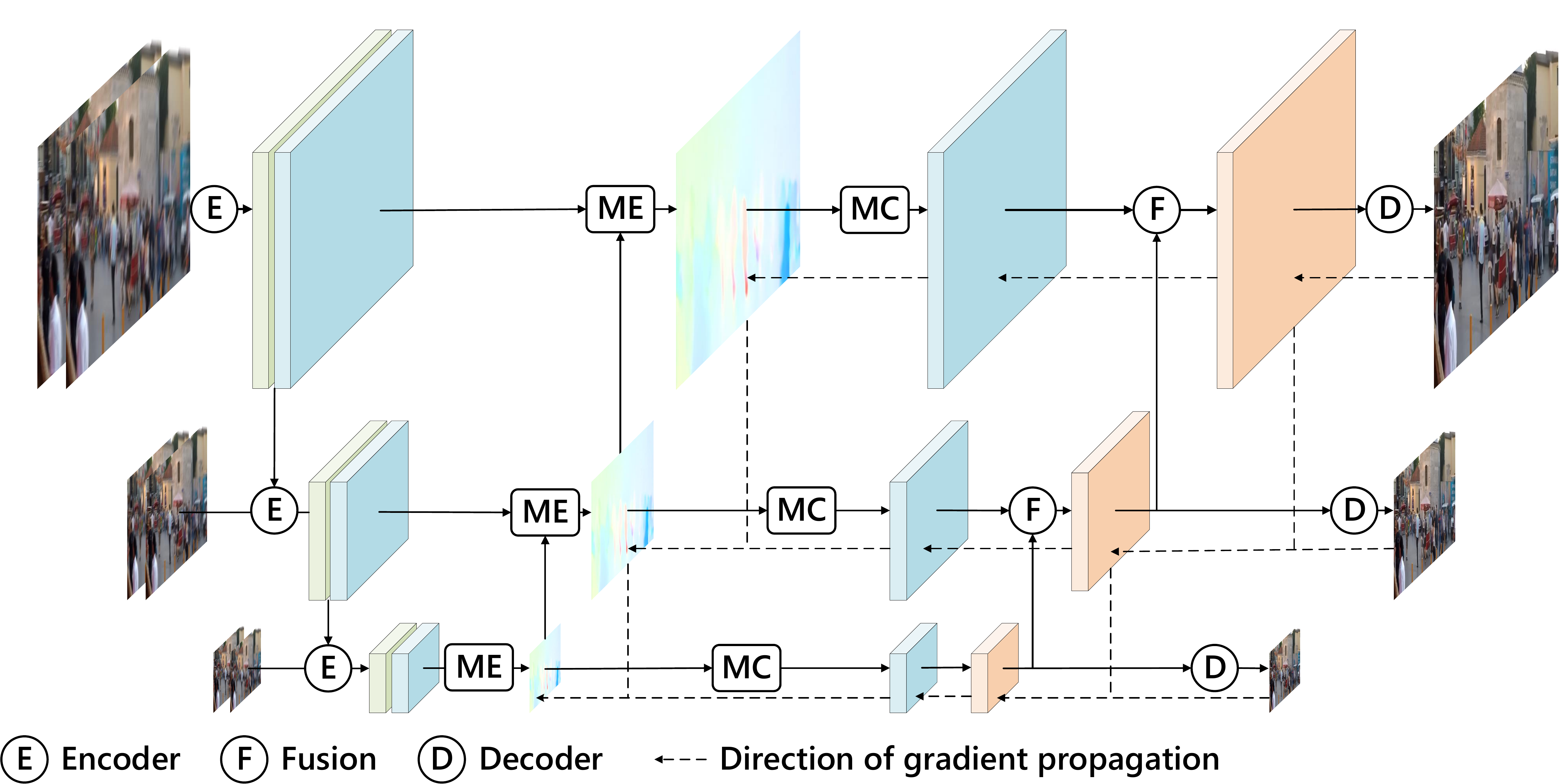}
  \caption{Pyramid Alignment}
  \label{fig:PA}
\end{subfigure}
\caption{Comparisons of the different alignment structures for video deblurring. (a) An Example that performs alignment at a single scale, such as EDVR~\cite{wang2019edvr}. (b) The proposed pyramid alignment. The dash line shows the direction of gradient flow of  the supervision for motion at different scales.}
\label{fig:intro}
\vspace{-3mm}
\end{figure*}

Although these methods have achieved advanced performance (see Fig.~\ref{fig:GoPro}), there are still two problems remaining to be addressed: (1) The coarse-to-fine structure is adopted both in the alignment and the reconstruction module, which introduces additional parameters, computational cost, and unavoidably increases the training difficulty. (2) Misalignment caused by lack of adequate supervision. Errors in the estimated motion may introduce artifacts to the aligned frame or feature map and adversely affect the performance of the reconstruction. Meanwhile, motion estimation from coarse to fine is necessary to handle large and complex motions. As shown in Fig~\ref{fig:SLA}, the supervision of motion estimations from a low scale comes from high-scale motion estimations in the traditional single-level alignment method, and the motion estimated in low-scale is weakly supervised. Therefore, we naturally have a question: \textit{how well the alignment  at low-scale could be utilized?} If there are errors in the alignment of the lower layer of the pyramid, the errors in coarser levels will accumulate and affect the final alignment due to the lack of supervision. Consequently, accumulated misalignment causes undesirable artifacts in restored images.

To address the above problems, we propose a novel pyramid framework for video deblurring (see Fig. ~\ref{fig:PA}). Specifically, a cascade CNN is first designed to extract multi-scale features from blurry frames. As downsampling in the feature pyramid construction may introduce blurry boundaries that adversely affect the alignment performance, a Structure-to-Detail Downsampling strategy is presented for boundary sharpening. Then, a novel Cascade Guided Deformable Alignment (CGDA) method is proposed to align the pyramid features of the target and neighboring frames from multiple scales. Instead of directly learning the offsets, the proposed CGDA uses lower-scale learned offset to guide the higher-scale offset estimation. Through this strategy, more accurate alignment can be achieved with more superficial network structures. After the alignment, the multi-scale decoders recover the multi-scale sharp images from features of aligned neighboring frames and the target frame. Finally, we introduce a multi-scale reconstruction loss to supervise the training dynamic to reduce the propagation of errors such that features can be better aligned for reconstruction. As shown in Fig. \ref{fig:GoPro}, PFAN makes a significant reduction in computational complexity and time complexity, whilst achieving the superior quality of the restored frame over existing methods for video deblurring.

Our contributions are summarized as follows:
\begin{itemize}
  \item A Pyramid Feature Alignment Network is proposed to exploit multi-scale information more effectively for video deblurring. The framework aligns the features in a coarse-to-fine manner, where reconstructions are cascaded refined via the incorporation of higher-resolution frame features. Through this strategy, alignment errors in coarse motion estimation due to factors such as occlusion and large motion can be significantly reduced, and error propagation can be prevented.

  \item A novel Cascade Guided Deformable Alignment (CGDA) is proposed to align features of neighboring frames with the target frame in a coarse-to-fine manner. CGDA uses learned motion information from lower scales to guide motion estimation at higher scales, which significantly improves the alignment accuracy while improving the training stability with reduced parameters.

  \item Our proposed method efficiently exploits the temporal and multi-scale information of blurry frames, and it reconstructs high-quality frames from blurry videos with higher computational efficiency. Extensive experiments demonstrate the proposed framework performs favorably against various state-of-the-art methods with reduced computational complexity.

\end{itemize}
\begin{figure*}
  \centering

  \includegraphics[width=0.95\textwidth]{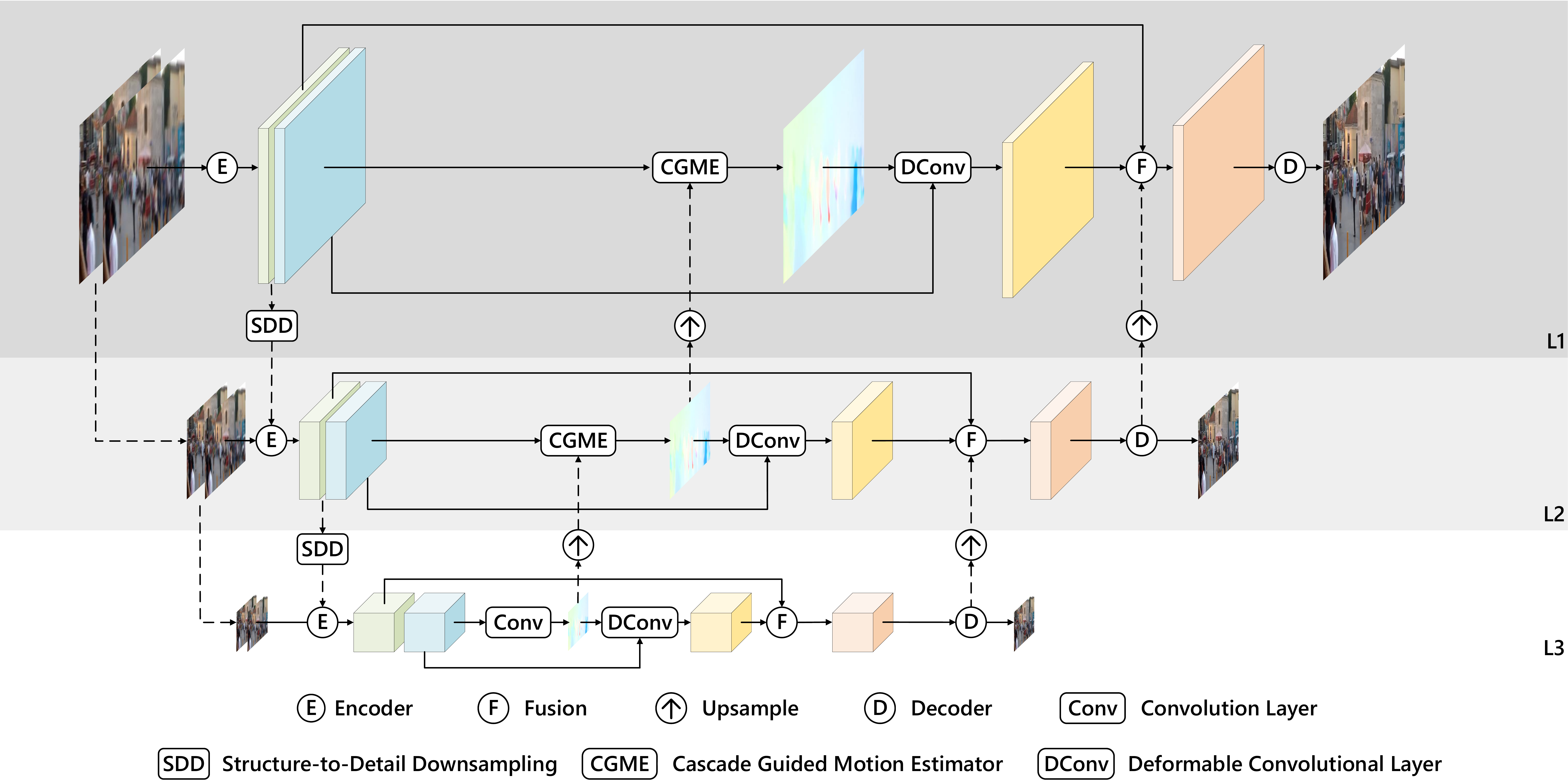}
  \caption{The proposed Pyramid Feature Alignment Network framework.}
  \label{fig:framework}
\vspace{-5mm}
\end{figure*}
\section{Related Work}
\subsection{Image Deblurring}
Traditional image deblurring methods make different assumptions to model blur characteristics~\cite{beck2009fast,dong2011image,gupta2010single, hirsch2011fast, joshi2009image}. However, they could not be generalized and used in practice due to the non-uniform blur kernels. The single-scale deblurring methods \cite{sun2015learning,su2017deep, shen2019human,zhang2020deblurring} tried to reconstruct high-resolution images directly, which usually leads to a time-consuming network with limited generalization. Muti-scale architectures for deblurring can progressively restore sharp images at each scale. Nah \emph{et al.}\cite{nah2017deep} pioneered a multi-scale CNN to restore latent images.  Tao \emph{et al.} \cite{tao2018scale} presented to a scale-recurrent ConvLSTM to aggregate the feature from coarse to fine. Gao \emph{et al.}\cite{gao2019dynamic} proposed a network with the strategy of selective parameter sharing and nested skip connections. Cho \emph{et al.} \cite{cho2021rethinking} presented a single U-Net-based architecture with significant modifications achieving advanced performance. Hu \emph{et al.} \cite{hu2021pyramid} designed a pyramid neural architecture to search hyper-parameters for deblurring automatically. Pyramid reconstruction networks have shown remarkable performance improvements in image deblurring. This structure is naturally adopted in video deblurring to handle spatially invariant blur.
\subsection{Video Deblurring}
Traditional methods model the blur as optical flow \cite{wulff2014modeling, hyun2015generalized, bar2007variational, lee2012video}. Hand-crafted priors show limited performance facing the temporally and spatially invariant blur. Some methods started to use CNNs to utilize information from neighboring frames. Some methods \cite{su2017deep, kim2018spatio, xue2019video, xiang2020deep} computed optical flow frames and used the aligned frames to enhance the target frame. Kim \emph{et al.} \cite{kim2018spatio} proposed a filter adaptive convolutional layer for the generated element-wise filters to feature transformation. Wang \emph{et al.} \cite{wang2019edvr} designed the pyramid, cascading, and deformable convolution module for better alignment. Pan \emph{et al.} \cite{pan2020cascaded} proposed a cascaded deblurring approach while utilizing temporal sharpness prior to reconstruction. Suin \emph{et al.} \cite{suin2021gated} introduced a factorized spatio-temporal attention and selected the key-frames instead of performing the explicit alignment. Li \emph{et al.} \cite{li2021arvo} proposed to adopt a pyramid correlation volume to learn the spatial correspondence among blurry frames.  Son \emph{et al.} \cite{son2021recurrent} presented blur-invariant motion estimation methods and a way to resolve motion estimation errors. However, these methods for information aggregation focused at a particular scale while ignoring to explore multi-scale information for better alignment and reconstruction. 

\section{The Proposed Method}
Given a $2N+1$ consecutive blurry frame sequence $\lbrace B_{t-N},\cdots,B_{t},\cdots,B_{t+N}\rbrace$ with $B_{t}$ as the target frame for deblurring, our task is to restore the sharp frame $S_{t}$. As shown in Fig.~\ref{fig:framework}, the proposed Pyramid Feature Alignment Network (PFAN) consists of three components: a feature extraction module, an alignment module and a reconstruction module. As the U-Net~\cite{ronneberger2015u} structure has shown superior performance in image deblurrring \cite{cho2021rethinking}, our network is also based on U-Net.

\subsection{Pyramid Feature Extractor}\label{subsection:MFE}
Under the coarse-to-fine principle to deal with spatially invariant blur, we present the pyramid feature extractor to construct a featured pyramid to explore multi-scale information for deblurring. Compared with the standard method to construct feature pyramid~\cite{lin2017feature}, using the image pyramid obtained by downsampling as an additional input achieved advanced performance, both in high-level tasks~\cite{chen2017person, pang2019efficient, nie2019enriched} and low-level tasks~\cite{guo2018hierarchical, cho2021rethinking}. Firstly, we extract features from blurry image pyramid. The feature extractor in the blurry pyramid can be expressed as follows:
\begin{equation}
    B_{t}^{k} = {Down}\{B_{t}^{k+1}\}, k = 2, 3
\end{equation}
\begin{equation}
    E_{t}^{k} = X_{k}\{B_{t}^{k}\}, k = 1, 2, 3
\end{equation}
where $B_{t}^{k}$ is the $t$-th blurry frame at level $k$, ${Down}$ denotes bilinear downsampling the image by a factor of 2, $X_{k}$ represents the feature extractor at level $k$ consisted of three convolution layers and ReLU layers, and $E_{t}^{k}$ is the extracted feature of $t$-th blurry frame at level $k$.

For general downsampling for feature pyramid construction, the boundaries between objects at low-scale feature maps appear hard to distinguish. This may mislead the motion estimation to mistakenly regard two objects as one object and thus incorrectly estimate this part of the motion. Therefore, in order to ensure sharp edges of features, we present a Structure-to-Detail Downsampling (SDD) strategy to downsize the feature map instead of directly downsampling the feature map with a convolution layer with a stride of 2. The proposed SDD consists of a max-pooling and a convolutional layer. Max-pooling enhances the boundaries between different objects visually. So we employ the max-pooling layer with 2 × 2 filters and stride 2 to downsample the feature map as the structure map. Since the max-pooling layer cannot restore all the details, we use a convolution layer with a stride of 2 to extract details to supplement the information of the structure map. The SDD is represented as follows:
\begin{equation}
  SDD(F_t^{k})=MP(F_t^{k})\ +\ Conv(F_t^{k}), \quad k=1,2
\end{equation}
where ${SDD}$ stands for the proposed Structure-to-Detail Downsampling function with a factor of 2, $MP$ denotes the max-pooling layer, $Conv$ denotes the convolution layer, $F_t^{k}$ denotes the feature of $t$-th frame at the $k$-th scale ($k=1,2$). After downsampling with SDD, we add a convolutional layer to increase the number of channels of the feature map. 

In addition to the downsized feature, we adopt the feature extracted from the downsampled blurry image as an additional input for the feature pyramid:
\begin{equation}
    F_{t}^{k}= \begin{cases}Encoder_{k}\left(E_{t}^{k}\right), & k=1 \\ Encoder_{k}\left(E_{t}^{k} ; {SDD}\left(F_{t}^{k-1}\right)\right), & k=2,3\end{cases}
\end{equation}
where $F_{t}^{k}$ is the encoded feature of the $t$-th blurry frame at level $k$,  $Enconder_{k}$ stands for the feature encoder of level $k$, which consists of six modified residual blocks~\cite{tao2018scale}.

\begin{figure}
  \centering
  \scalebox{0.95}{
  \includegraphics[width=1\linewidth]{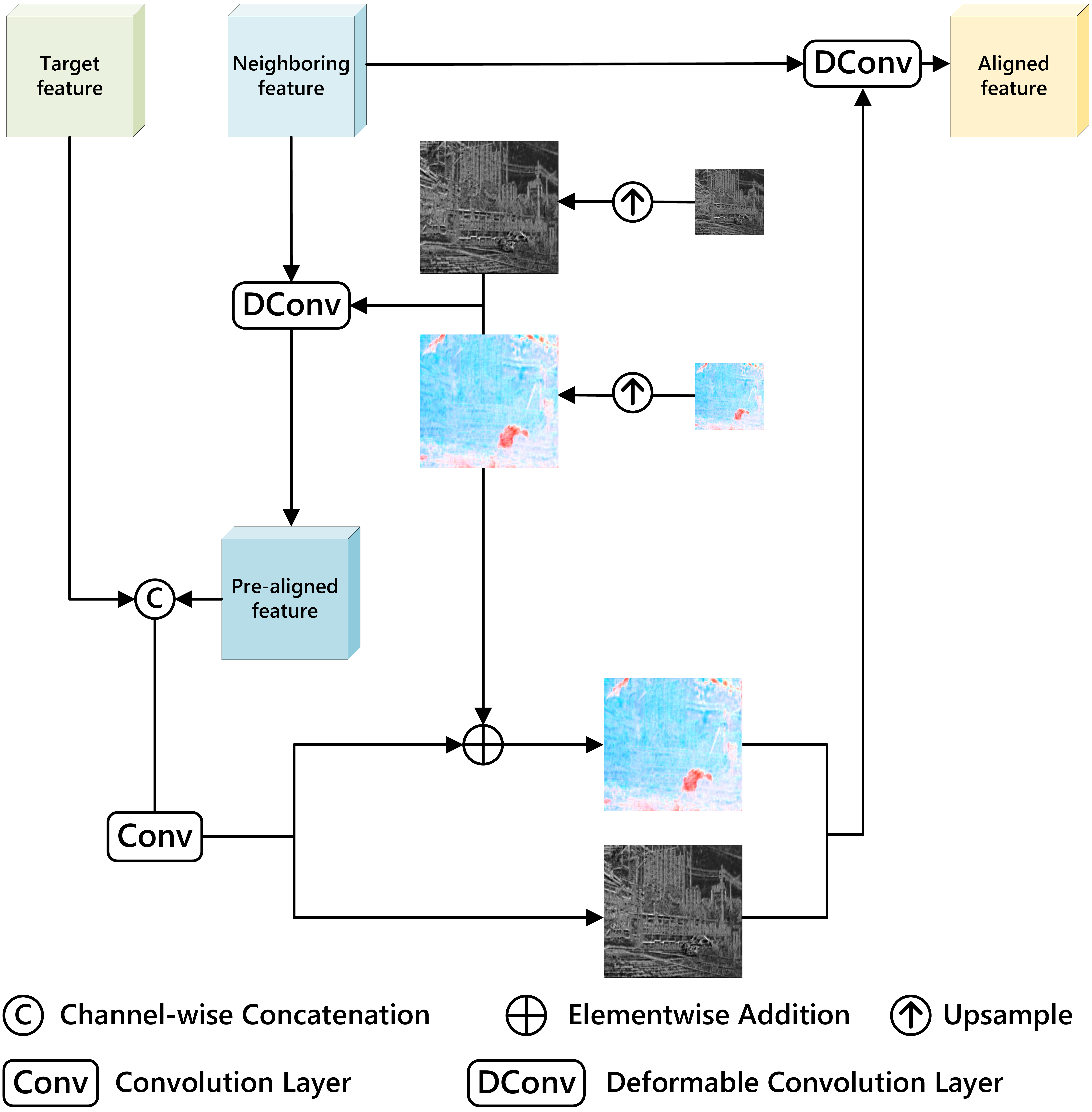}
  }
  \caption{The framework of Cascade Guided Deformable Alignment module.}

  \label{fig:CGCA}
\vspace{-4mm}
\end{figure}
\subsection{Cascade Guided Deformable Alignment}~\label{subsection:CGDAM}
Due to the existence of large and complex motion, accurate feature alignment is challenging for video deblurring. Using optical flow to align frames requires a complex and time-consuming structure. For example, the number of the parameters of the optical flow model, PWC-Net~\cite{sun2018pwc}, used in CDVD-TSP reaches 6.19M. However, Deformable Conventional Network (DCN) \cite{dai2017deformable} predicts the sampling locations (offset) and the weight of each sample. The deformable alignment shows its superiority for video enhancement owing to the diversity of offsets~\cite{chan2021understanding}.  However, training the deformable-based alignment module is hard in practice~\cite{chan2021understanding}. Meanwhile, since estimating large motion is challenging, increasing the number of DCN layers is necessary to expand the receptive field. Inspired by the coarse-to-fine strategy in classical optical flow estimation methods~\cite{horn1981determining, black1996robust, brox2004high, ranjan2017optical}, we propose Cascade Guided Deformable Alignment (CGDA), which only uses cascade three DCN based modules to align the multi-scale features. As shown in Fig. \ref{fig:CGCA}, instead of learning the offset directly, the proposed module utilizes the predicted low-scale offset to guide the higher-scale offset estimation.

Specifically, taking the feature of the target frame $F_{t}$ and the feature of the reference frame $F_{t-1}$ as an example, to get the aligned feature $\hat{F}_{t-1}^{k}$, their concatenation in the channel direction $c(F_{t}, F_{t-1})$ is used to generate the offset and corresponding modulation scalars through several convolutional layers. Firstly, CNN is directly performed at the lowest scale ($k=3$) to obtain the offset and mask:
\begin{align}
        &o_{t\rightarrow t-1}^{k} = \mathcal{G}_{o}\left(c(F_{t}^{k},F_{t-1}^{k})\right),\\
        &m_{t\rightarrow t-1}^{k} = \sigma\left(\mathcal{G}_{m}\left(c(F_{t}^{k}, F_{t-1}^{k})\right)\right),
\end{align}
where $o_{t\rightarrow t-1}^{k}$ and  $m_{t\rightarrow t-1}^{k}$ denote the DCN offset and the modulated mask between frame $t$ and $t-1$ at the $k$-th scale, $\mathcal{G}_{o}$ and $\mathcal{G}_{m}$ denote a group of convolution layers to generate the offset and mask, $\sigma$ denotes the sigmoid function. Therefore, the aligned feature $\hat{F}_{t-1}^{k}$ is calculated as:
\begin{equation}
  \hat{F}_{t-1}^{k} = {DConv}\left(F_{t-1}^{k}; o_{t\rightarrow t-1}^{k}, m_{t\rightarrow t-1}^{k}\right),
\end{equation}
where ${DConv}$ denotes a deformable convolution layer.

As deformable alignment can be formulated as a combination of feature-level flow-warping and convolution, the offset can  represent the motion information at each position, like optical flow~\cite{chan2021understanding}. The upsampled offset $\tilde{o}_{t\rightarrow t-1}^{k}$ and mask $\tilde{m}_{t\rightarrow t-1}^{k}$ for pre-alignment are calculated as:
\begin{align}
  &\tilde{o}_{t\rightarrow t-1}^{k} = {up}\left(o_{t\rightarrow t-1}^{k+1}\right) \times 2, \\
  &\tilde{m}_{t\rightarrow t-1}^{k} = {up}\left(m_{t\rightarrow t-1}^{k+1}\right),
\end{align}
where ${up}$ denotes the bilinear upsampling by the scale of 2. Then we use the upsampled offset $\tilde{o}_{t\rightarrow t-1}^{k}$ and mask $\tilde{m}_{t\rightarrow t-1}^{k}$  ($k=1, 2$) to pre-align the feature, and the pre-aligned $k$-th scale feature $\bar{F}_{t-1}^{k}$ becomes:
\begin{equation}
    \bar{F}_{t-1}^{k} = {DConv}\left(F_{t-1}^{k}; \tilde{o}_{t\rightarrow t-1}^{k}, \tilde{m}_{t\rightarrow t-1}^{k}\right).
\end{equation}

The concatenated features $c(F_{t}^{k}, \bar{F}_{t-1}^{k})$ are used to compute the offset ${o}_{t\rightarrow t-1}^{k}$ and modulation mask ${m}_{t\rightarrow t-1}^{k}$ between the pre-aligned feature $\bar{F}_{t-1}^{k}$ and target feature ${F}_{t}^{k}$ at the scale $k$. Instead of directly computing the DCN offset, we compute the final offset ${o}_{t\rightarrow t-1}^{k}$ and mask ${m}_{t\rightarrow t-1}^{k}$ as:
\begin{align}
  &{o}_{t\rightarrow t-1}^{k} = \tilde{o}_{t\rightarrow t-1}^{k} + \mathcal{G}_{o}\left(c(F_{t}^{k}, \bar{F}_{t-1}^{k})\right),\\
  &{m}_{t\rightarrow t-1}^{k} = \sigma\left(\mathcal{G}_{m}\left(c(F_{t}^{k}, \bar{F}_{t-1}^{k})\right)\right),
\end{align}
where $\mathcal{G}_{o}\left(c(F_{t}^{k})\right)$ denotes the residual offset which is used to refine the upsampled offset and correct errors in ${o}_{t\rightarrow t-1}^{k}$. Then DCN is applied to warp the feature $F_{t-1}^k$ to obtain the aglined feature $\hat{F}_{t-1}^{k}$ at scale $k$ as:
\begin{equation}
    \hat{F}_{t-1}^{k} = {DConv}\left({F}_{t-1}^{k}; {o}_{t\rightarrow t-1}^{k}, {m}_{t\rightarrow t-1}^{k}\right),
\end{equation}

In summary, our proposed CGCA implements the alignment of the pyramid feature with a simple structure. Each layer in CGCA has a simplified task compared with the full motion estimation. The alignment module of each layer only needs to estimate the small motion update of the existing offset. Furthermore, to avoid the accumulation of the misalignment at a higher level when there exist errors in the motion estimation at the coarse-scale, we design the Pyramid Reconstruction strategy described in the following section to prevent errors from accumulating in motion estimation.
\subsection{Pyramid Reconstruction}\label{subsection:PFRM}
Modeling and exploiting the temporal-spatial relation of feature sequences is useful. For reconstruction of sharp frames, a temporal-spatial attention block \cite{wang2019edvr} is introduced to our model to fuse pyramid features of the target frame and the aligned neighboring feature is calculated as: 
\begin{equation}
  U^{k} = {TSA}_{k}(\ \hat{F}_{t-N}^{k}, \cdots, {F}_{t}^{k}, \cdots, \hat{F}_{t+N}^{k}\ ),
\end{equation}
where ${TSA}_{k}$ denotes the temporal-spatial attention fusion at the $k$-th scale, $U^{k}$ stands for the fused feature at the scale $k$, $\hat{F}_{t-N}^{k}$ denotes the aligned feature of the frame $t-N$ at the $k$-th scale, and ${F}_{t}^{k}$ denotes the feature of the target frame at the $k$-th scale.

After decoding the fused feature, we reconstruct the image at each scale:
\begin{equation}
  \hat{S^{k}}= \begin{cases}Conv\left({D}_{k}\left({U}^{k} ; up\{{D}_{k-1}({U}^{k-1})\}\right)\right), & k=2, 3 \\ Conv\left({D}_{k}\left({U}^{k}\right)\right), & k=1\end{cases}
\end{equation}
where ${D}_{k}$ is the feature decoder block at the level $k$ consisting of six modified residual
blocks~\cite{tao2018scale}, $Conv$ means a single convolution layer to generate the sharp image, and $up$ means the bilinear upsampling by with the factor of 2.
The recovered sharp image pyramid is then utilized to supervise the training dynamic.
\begin{table*}[t]
  \centering
  \caption{Quantitative comparisons against deblurring methods on the GoPro ~\cite{nah2017deep} and DVD~\cite{su2017deep} test datasets. The best and second performance are highlighted with \textbf{bold} and \underline{underline}, respectively.
  }
  \vspace{-2mm}

    \begin{tabular}{cccccccc}
    \toprule
    \multirow{2}[4]{*}{Methods} & \multirow{2}[4]{*}{frame} & \multicolumn{2}{c}{GoPro } & \multicolumn{2}{c}{DVD} & \multirow{2}[4]{*}{Parameters ($\times 10^{6}$)} & \multirow{2}[4]{*}{Runtime (s)} \\
\cmidrule{3-6}          &       & PSNR (dB)  & SSIM  & PSNR (dB)  & SSIM  &       &  \\
    \midrule
    Tao et al.~\cite{tao2018scale} & 1     & 30.29 & 0.9014 & 29.98 & 0.8842 & 8.06      & 2.32  \\
    Su et al.~\cite{su2017deep} & 5     & 27.31 & 0.8255 & 30.01 & 0.8877 & 15.30 & 6.14 \\
    Nah et al.~\cite{nah2017deep} & 1     & 29.97 & 0.8947 & - & - & 11.71 & 13.82 \\
    EDVR~\cite{wang2019edvr} & 5     & 26.83 & 0.8426 & 28.51 & 0.8637 & 23.60 & 0.32 \\
    STFAN~\cite{zhou2019spatio} & 2     & 28.59 & 0.8608 & 31.15 & 0.9049 & \textbf{5.37} & \textbf{0.15} \\
    TSP~\cite{pan2020cascaded} & 5     & 31.67 & 0.9279 & 32.13 & 0.9268 & 16.19 & 3.01 \\
    UHDVD~\cite{deng2021multi} & 2     & 31.33 & 0.921 & 32.19 & 0.937 & -     & - \\
    PVDNet~\cite{son2021recurrent} & 3     & 31.98 & 0.928 & 32.31 & 0.926 & 23.4 & 0.55 \\ 
    AVRo~\cite{li2021arvo} & 5     & - & - & 32.80 & 0.9352 & -     & - \\
    GSTAN~\cite{suin2021gated} & 5     & 32.10  & \textbf{0.96}  & 32.53  & 0.9468  & -     & - \\
    PFAN (Ours)  & 3     & \underline{32.74} & 0.9459 & \underline{33.01} & \underline{0.9478} & \underline{7.94} & \underline{0.21} \\
    PFAN+ (Ours)   & 5     & \textbf{32.89} & \underline{0.9469} & \textbf{33.21} & \textbf{0.9489} & 8.09 & 0.46 \\
    \bottomrule
    \end{tabular}%
  \label{tab:main}%
  \vspace{-3mm}

\end{table*}%
\subsection{Multi-scale Supervision}\label{subsection:loss}
The accuracy of motion estimation greatly affects the performance of video deblurring. Training the alignment module and reconstruction module in an end-to-end manner helps models learn task-oriented motion representation~\cite{xue2019video}. Therefore, instead of explicitly supervising the alignment, existing methods use reconstruction loss to supervise motion estimation indirectly (see Fig.~\ref{fig:SLA}). To handle large and complex motion, we apply the coarse-to-fine strategy for alignment. The previous methods have only weak supervision on low-scale motion estimation. However, it is challenging to correct the errors at the low-scale motion estimation when estimating the motion at the higher-scale. The accumulation of errors can cause severe misalignment between features. Furthermore, it may cause serious artifacts, resulting in performance degradation. Therefore, we propose to supervise the network at each scale to strengthen supervision for the low-scale motion estimation (see Fig.~\ref{fig:PA}). By comparing the gradient flow in the single-scale alignment and pyramid alignment, it shows that our proposed framework can better supervise the model for alignment. In Section \ref{section:exp}, we will demonstrate the ability of our method to reduce errors and correct the existing errors in motion estimation.

The multi-scale content loss $L_{cont}$ is defined as follows:
\begin{equation}
  L_{cont} = \sum_{k=1}^{K} {\dfrac{\mathcal{W}^{k}}{c^{k}w^{k}h^{k} } \parallel {\hat{S}^{k}} - S^{k}\parallel_{1}},
\end{equation}
where ${S}^{k}$ refers to the ground truth of the sharp frame. The loss at each scale is normalized by the number of channels $c_{k}$, width $w_{k}$, and  height $h_{k}$.
$\{\mathcal{W}^{k}\}$ are the weights for each scale. We empirically set $\mathcal{W}^{k}$ = 1.0. 

The restoration of high-frequency components is significant for deblurring, so we also adopted the frequency reconstruction loss function \cite{cho2021rethinking}:
\begin{equation}
  L_{FR} = \sum_{k=1}^{K} {\dfrac{1}{c^{k}w^{k}h^{k} } \parallel \mathcal{F}({\hat{S}}^{k}) - \mathcal{F}(S^{k})\parallel_{1}},
\end{equation}
where $\mathcal{F}$ denotes the fast Fourier transform (FFT) that transfers image signal to the frequency domain. The final loss function for training our network is calculated as follows:

\begin{equation}
  L = L_{1} + \lambda L_{FR},
\end{equation}
where $\lambda$ is the weight of frequency reconstruction loss.

\section{Experiments}

\subsection{Experiment Setup}
\textbf{Implementation Details.} In the training process, we use the Adam optimizer~\cite{kingma2014adam} with parameters $\beta_{1}$ = 0.9, $\beta_{2}$ = 0.999, and $\epsilon$ = $10^{-8}$. The minibatch size is set to be 3 per GPU. The learning rate is initialized to be $10^{-4}$ and decreases to half after every 300 epochs. We set the $L_{FR}$ weight $\lambda$ = 0.1.

\textbf{Datasets.} We use the GoPro \cite{nah2017deep} and DVD\cite{su2017deep} datasets for training the models under the formal training and testing settings of previous state-of-the-art methods. The GoPro and DVD have 33 and 71 videos with 3,214 and 6,708 blurry-sharp pairs, respectively. We randomly sample four images and randomly crop them with the size of $ 256 \times 256$ during training. For data augmentation, each patch is horizontally flipped with a probability of 0.5. 
  
\subsection{Comparisons with State-of-the-Art Methods}
To evaluate the performance of the proposed method,
we compare it against state-of-the-art methods~\cite{tao2018scale,su2017deep,nah2017deep,wang2019edvr,zhou2019spatio, pan2020cascaded, li2021arvo,suin2021gated,deng2021multi,son2021recurrent}. To evaluate the quality of each restored image on the datasets, we use PSNR and SSIM as the evaluation metrics. Considering the trade-off between the computational complexity and deblurring performance, we evaluate the performance of our framework with a different number of blurry frames for reconstruction, PFAN: 3 frames; PFAN+: 5 frames. In addition, we report the parameters and running time of various open-sourced methods on the same GTX1080 that is used to train our method.

\textbf{Quantitative Comparisons:}
The quantitative results on the GoPro ~\cite{nah2017deep} and DVD~\cite{su2017deep} test datasets are reported in Table \ref{tab:main}. The proposed method performs favorably against the state-of-the-art methods in terms of PSNR and SSIM.

\begin{figure*}[t!]
  \centering
  
  \setlength\tabcolsep{1.5pt}
  \scalebox{0.94}{
  \begin{tabular}{cccccc}
  \includegraphics[width=0.16\textwidth]{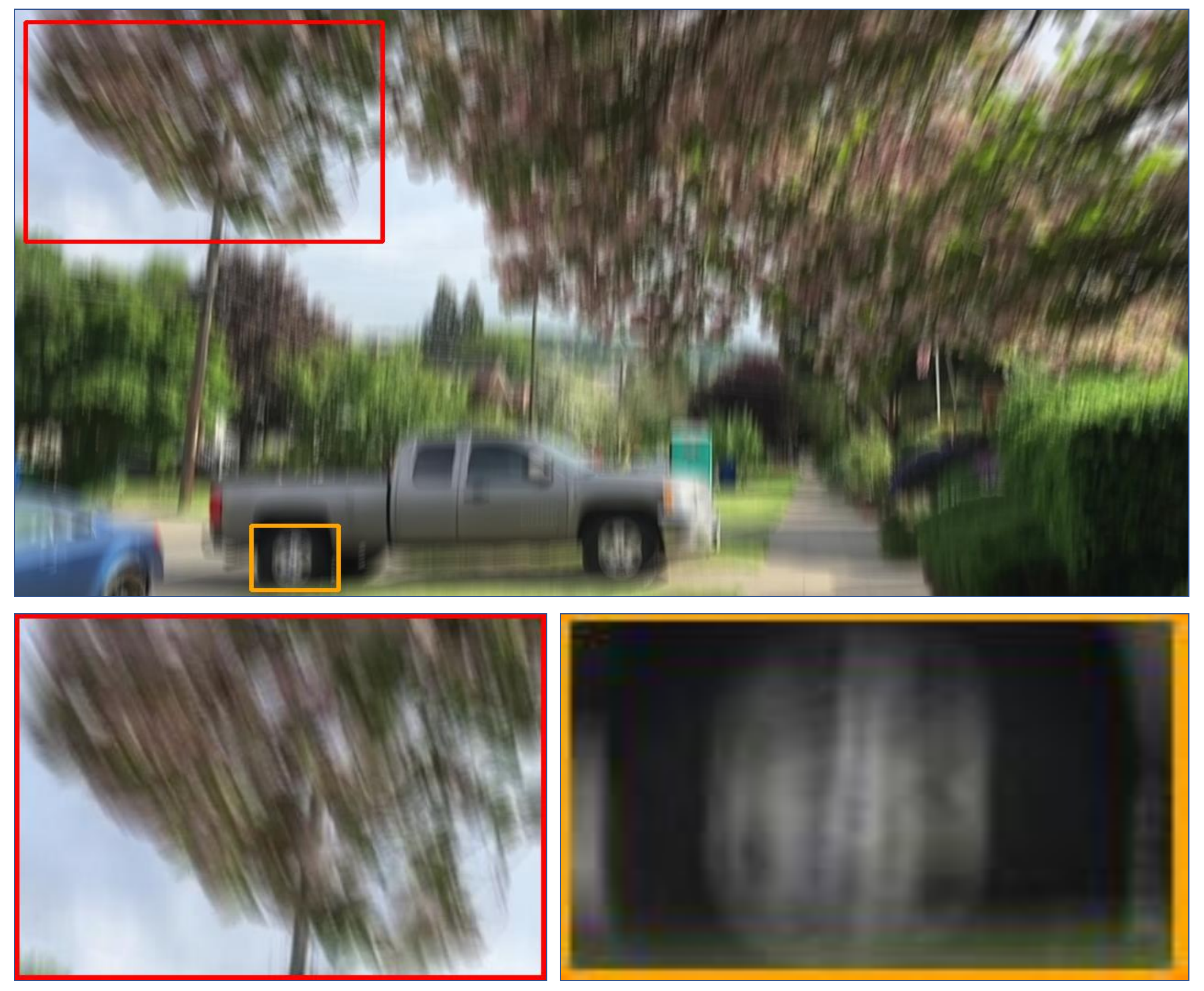}&
  \includegraphics[width=0.16\textwidth]{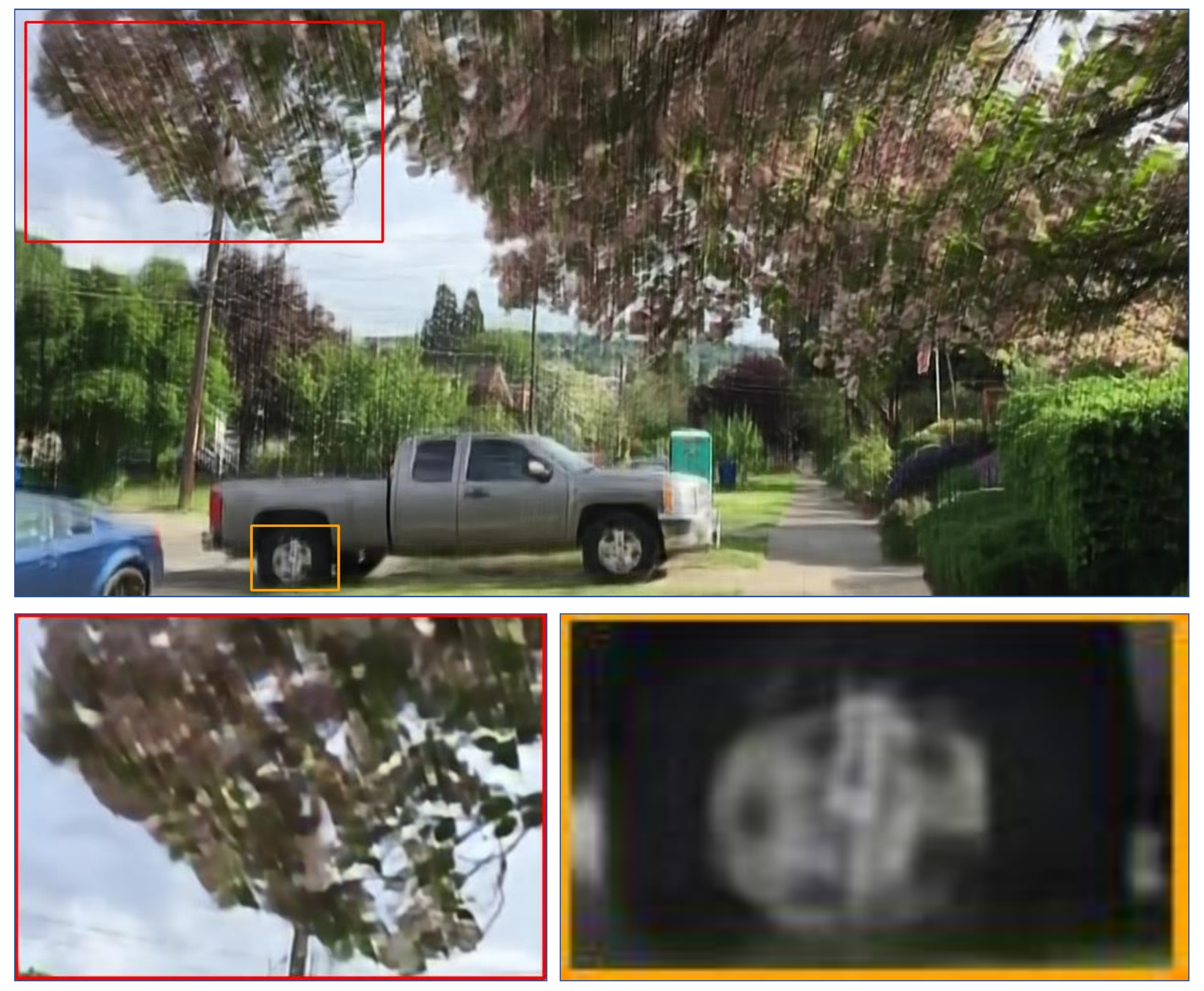}&
  \includegraphics[width=0.16\textwidth]{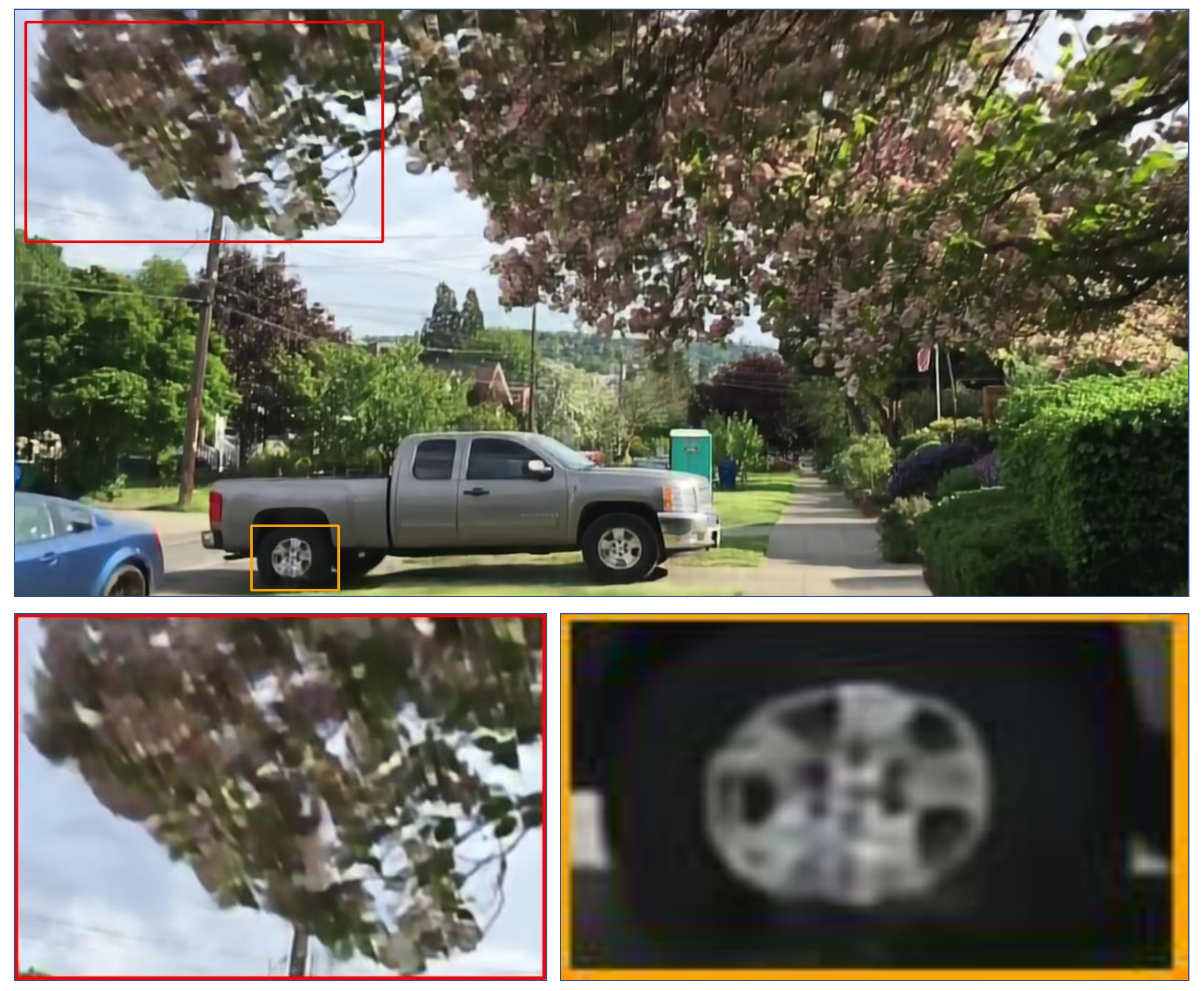}&
  \includegraphics[width=0.16\textwidth]{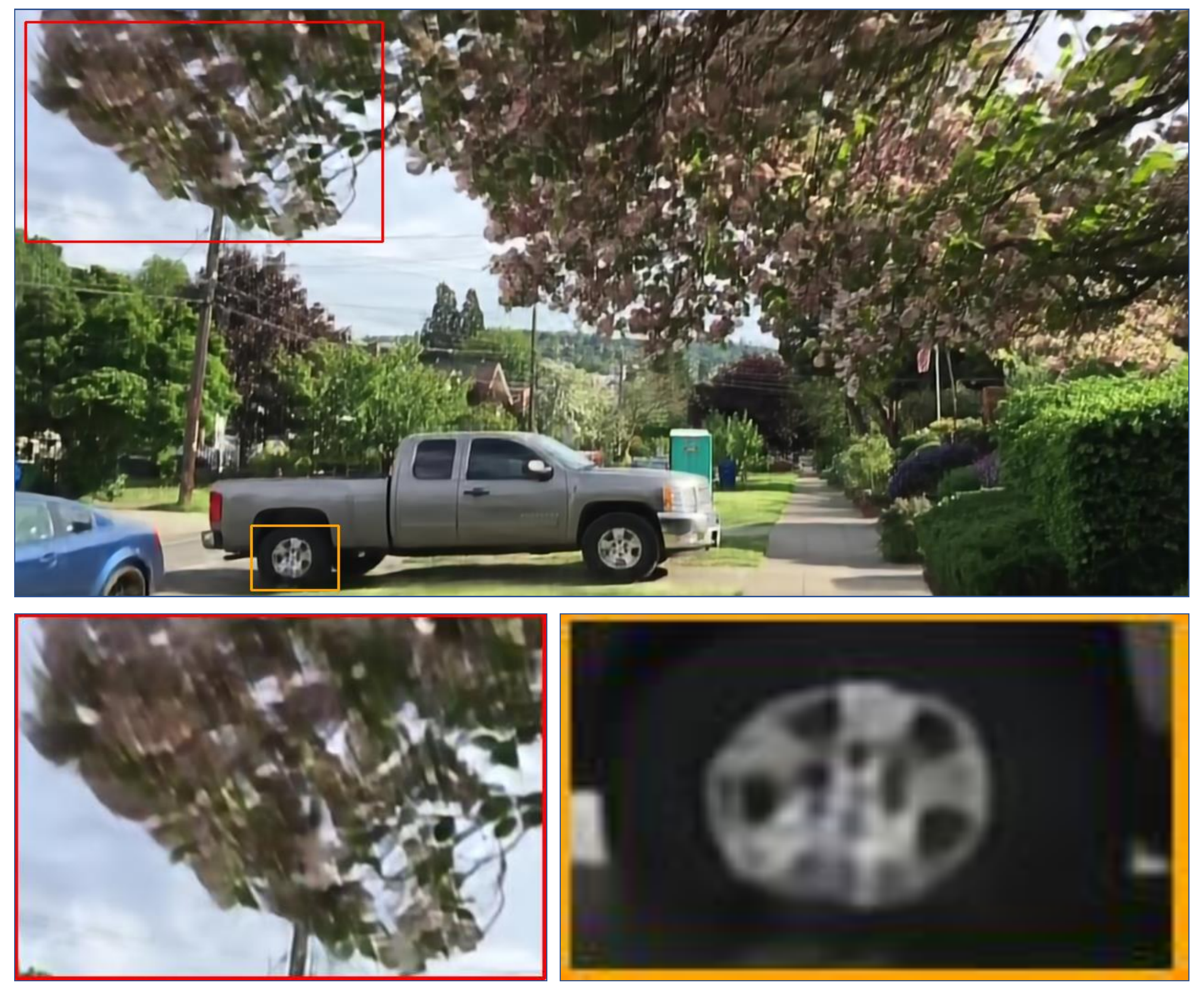}&
  \includegraphics[width=0.16\textwidth]{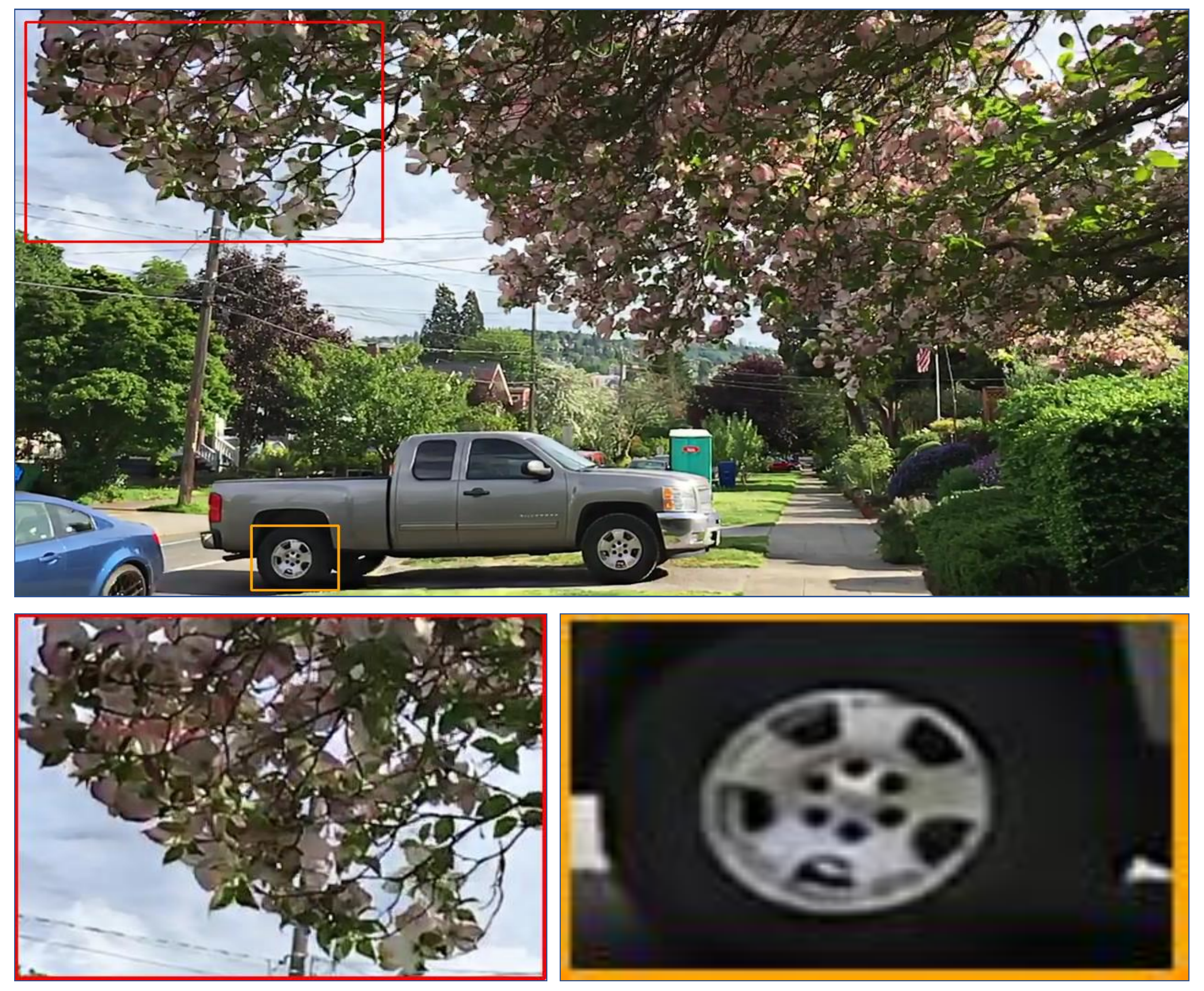}&
  \includegraphics[width=0.16\textwidth]{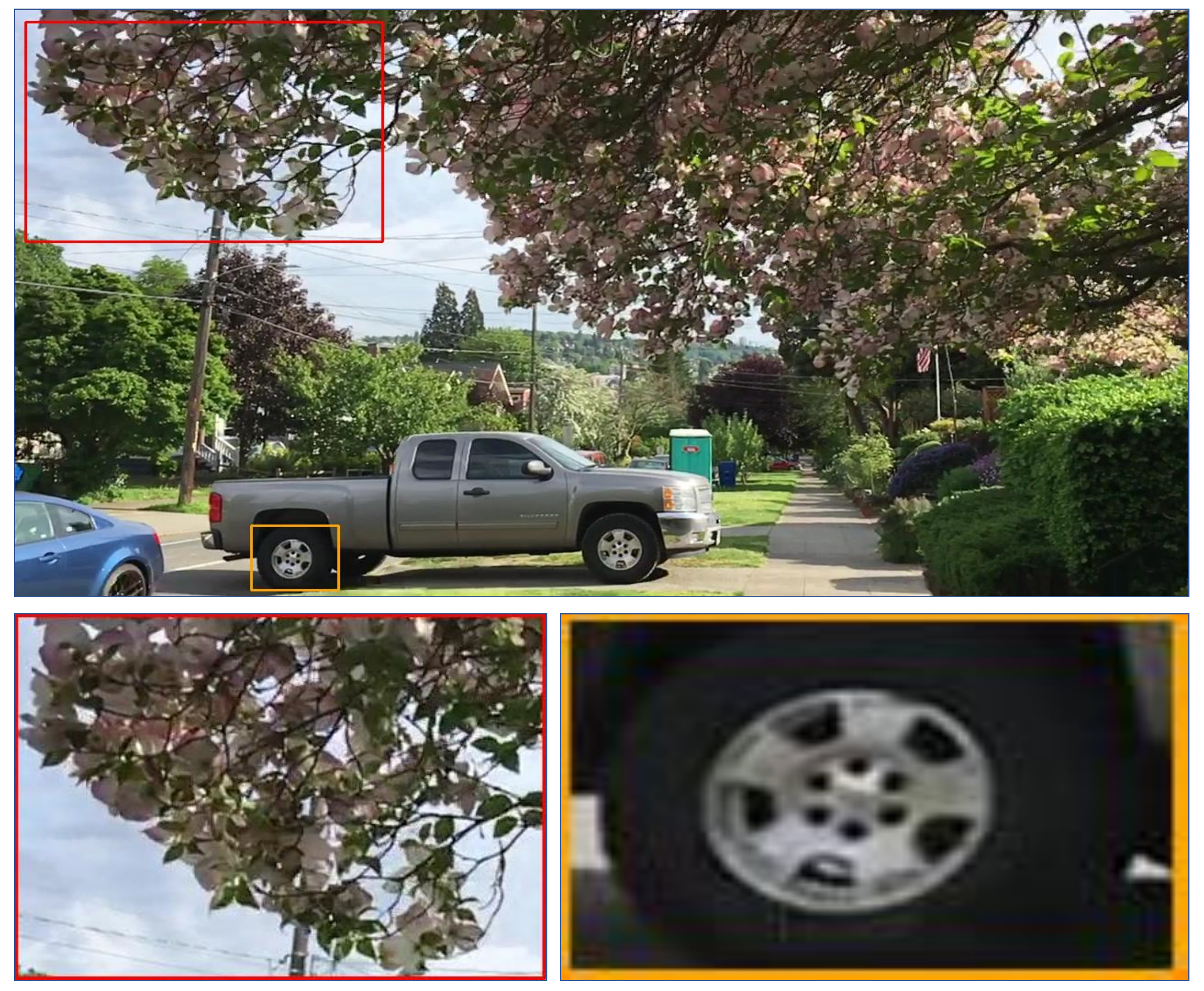}\\
  
  \includegraphics[width=0.16\textwidth]{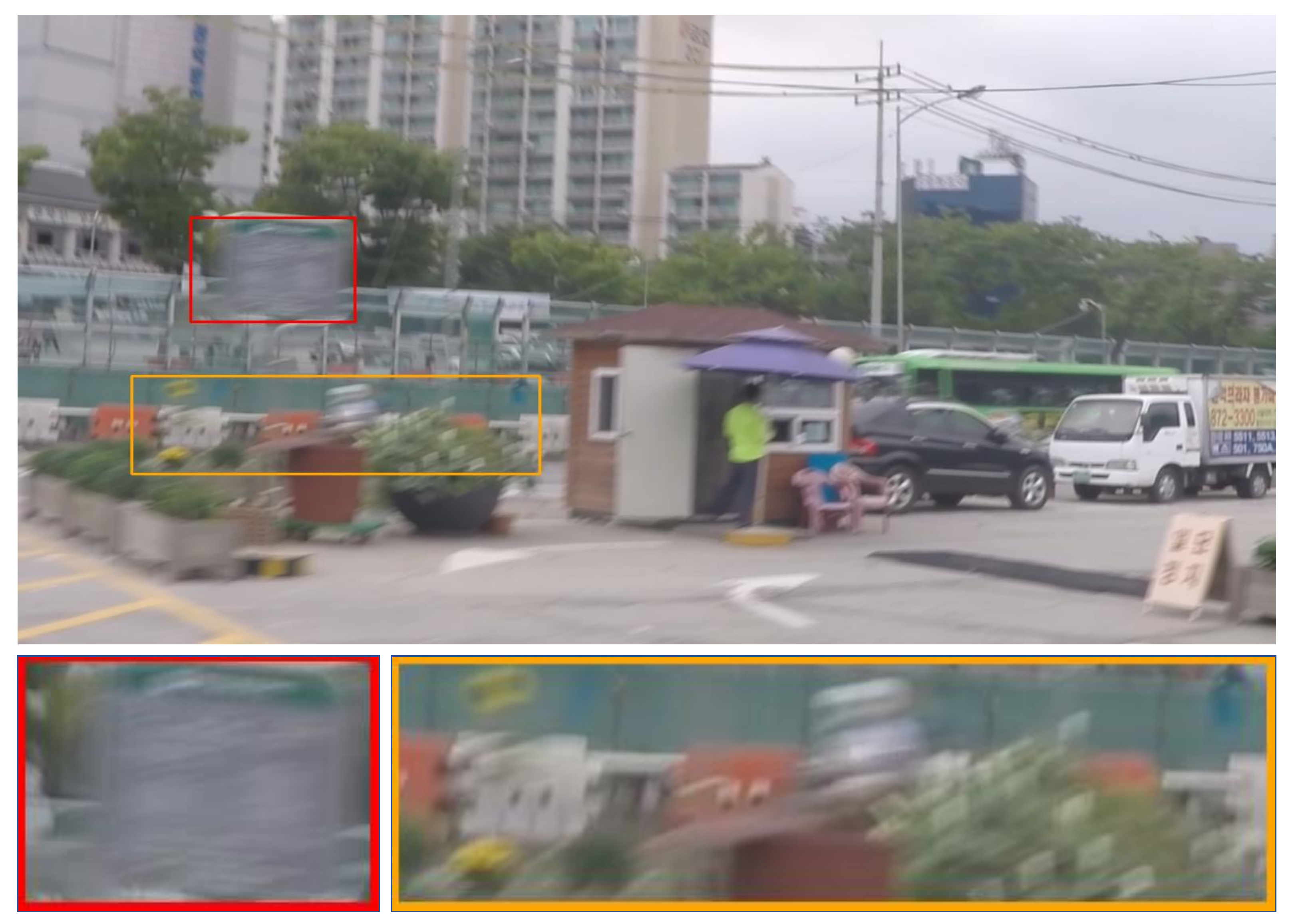}&
  \includegraphics[width=0.16\textwidth]{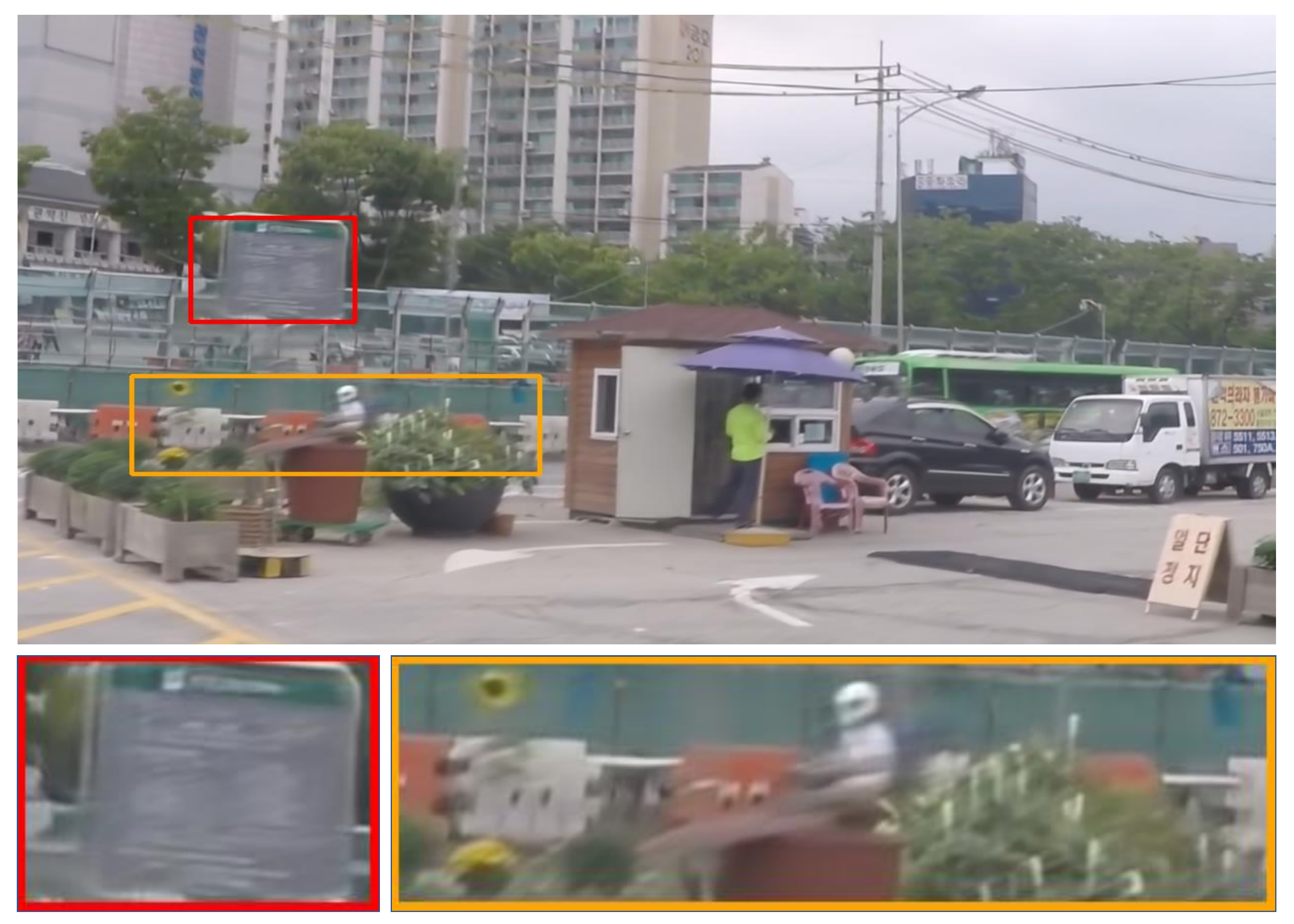}&
  \includegraphics[width=0.16\textwidth]{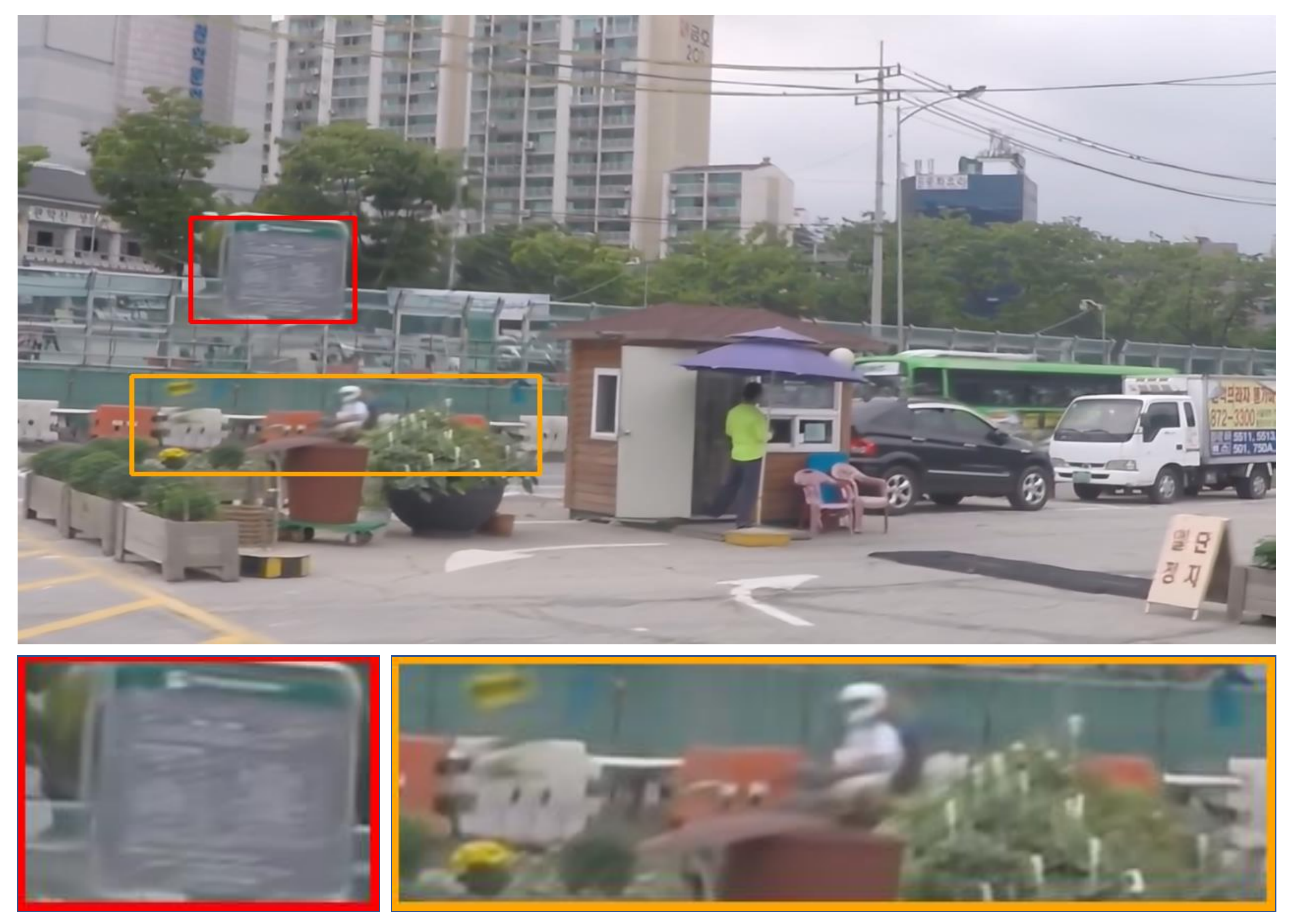}&
  \includegraphics[width=0.16\textwidth]{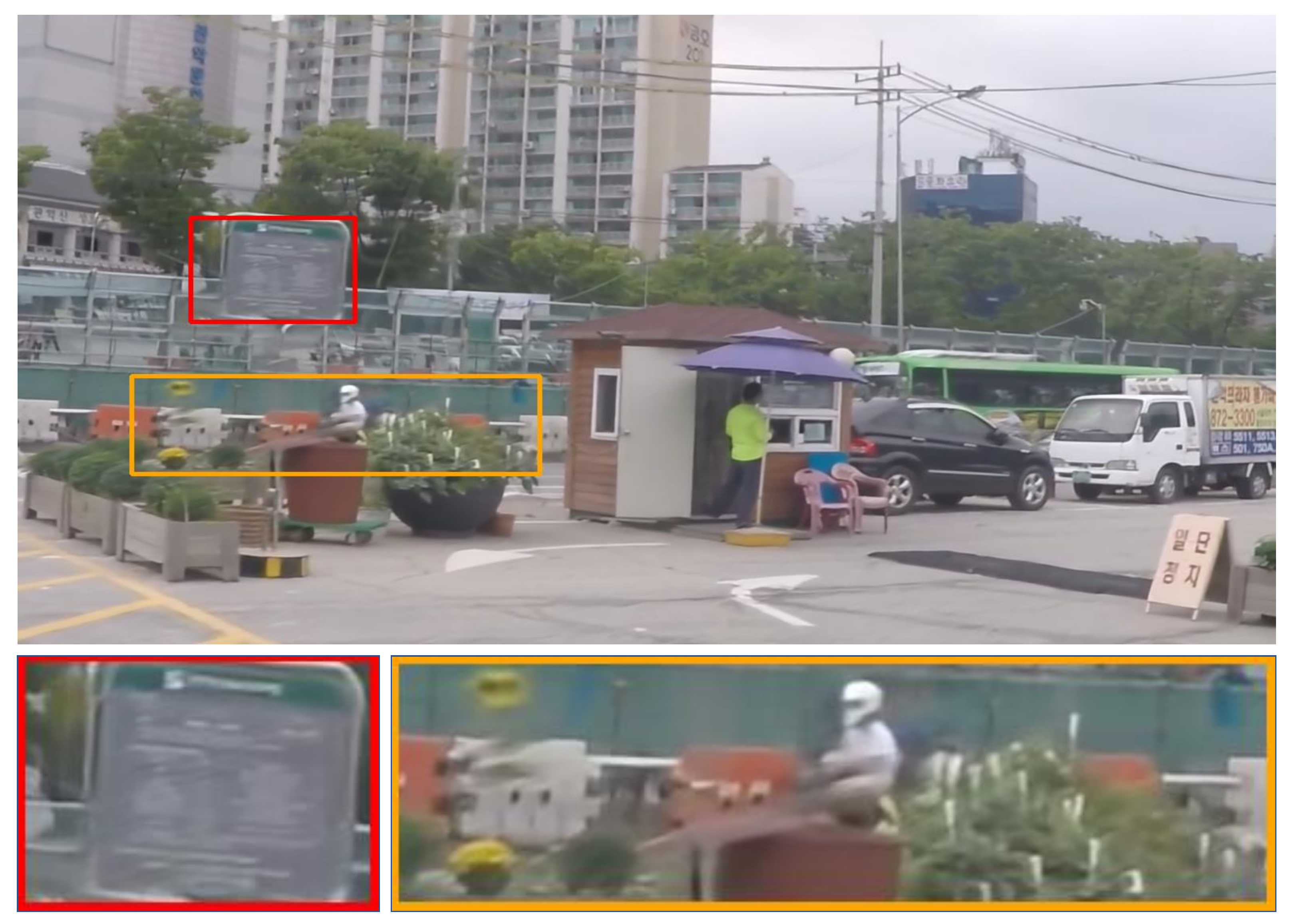}&
  \includegraphics[width=0.16\textwidth]{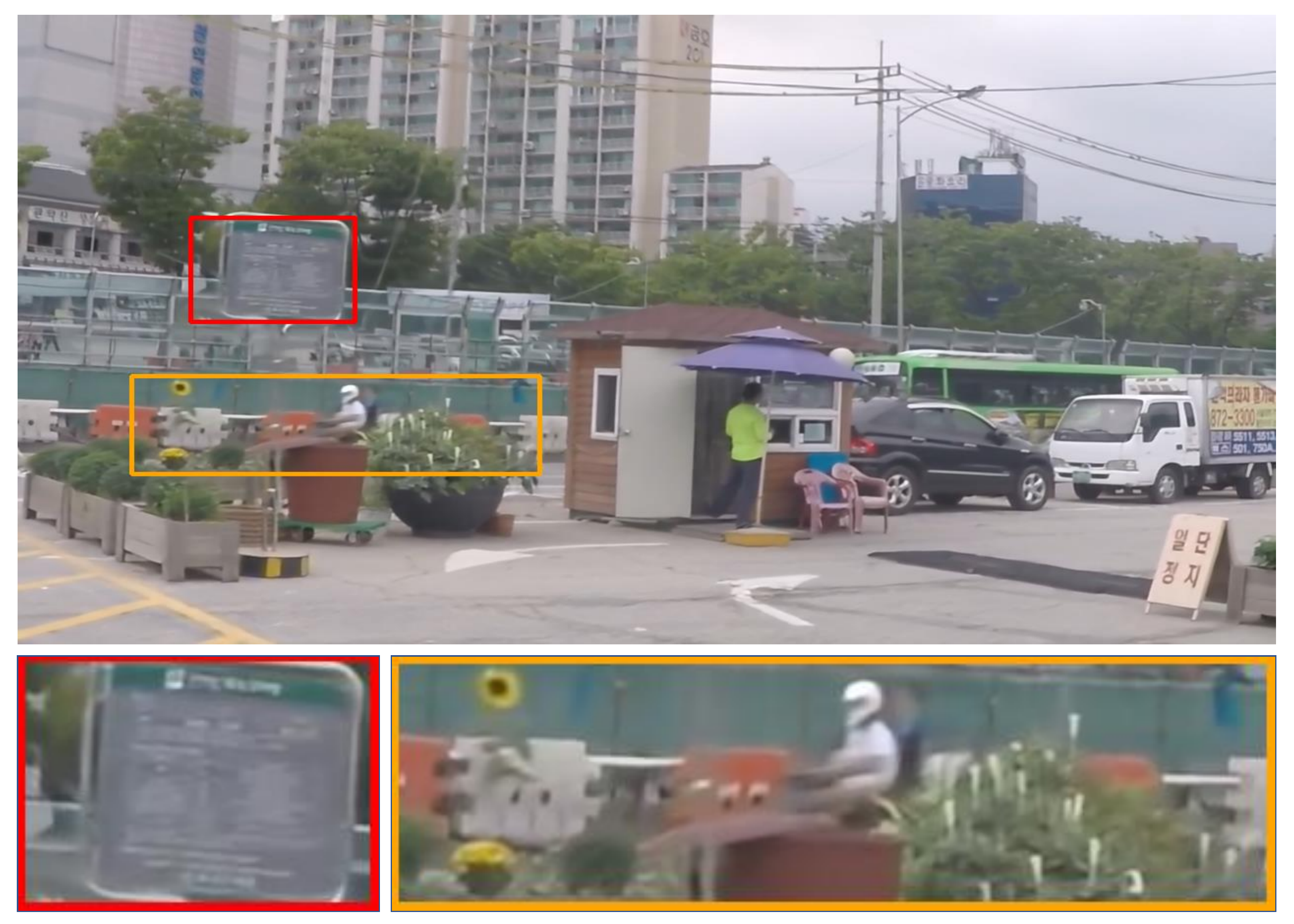}&
  \includegraphics[width=0.16\textwidth]{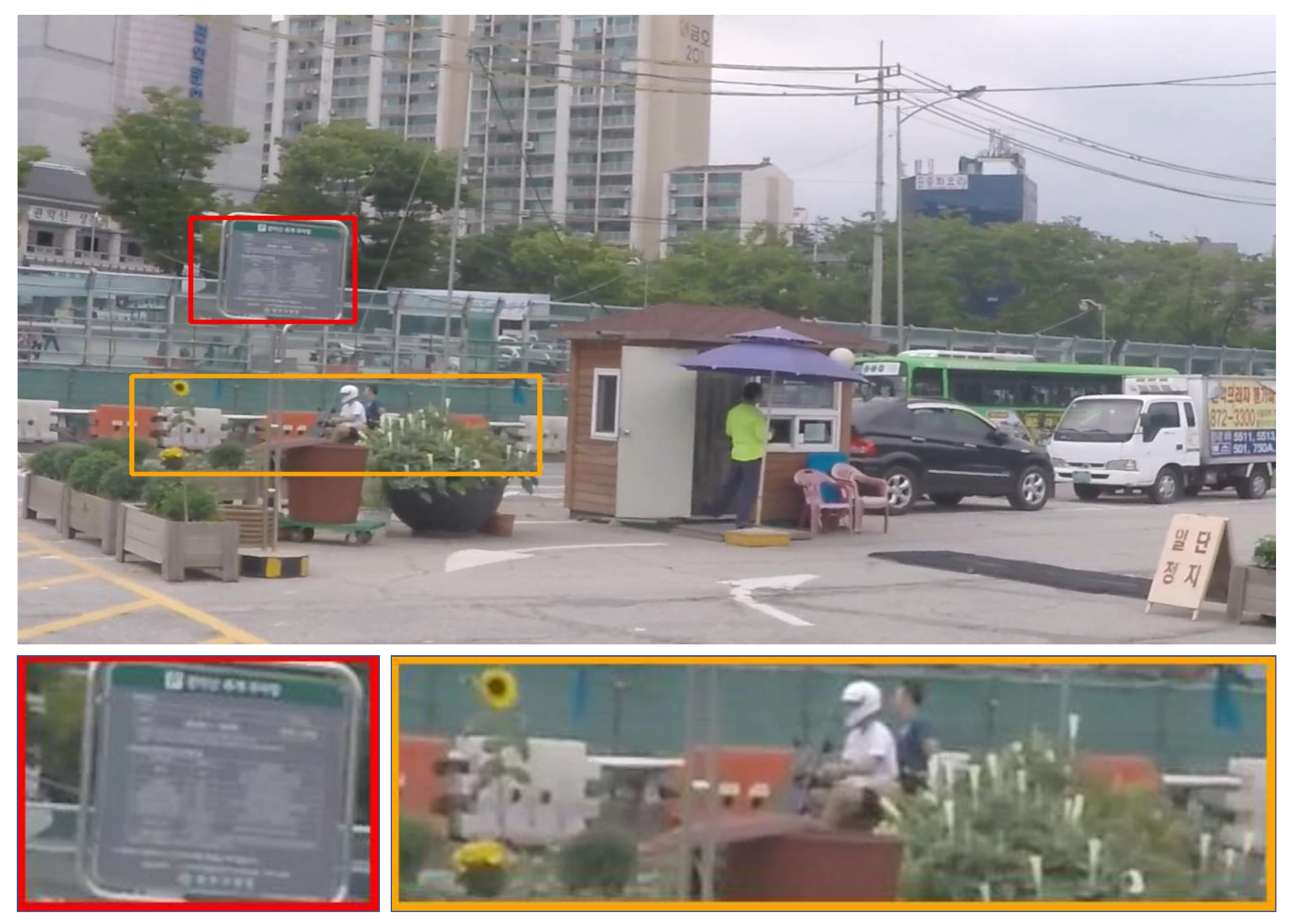}\\
  
  (a) Input & (b) STFAN~\cite{zhou2019spatio} & (c) TSP~\cite{pan2020cascaded} & (d) PVDNet~\cite{son2021recurrent} & (e) PFAN (Ours) & (f) Ground Truth
  
  \end{tabular}
  }
  \vspace{-2mm}
  
  \caption{Qualitative comparisons on the DVD~\cite{su2017deep} and GoPro ~\cite{nah2017deep} datasets. Our model benefits from the efficient and accurate alignment in feature space and makes more effective use of the information of neighboring frames to complete the restoration of sharp regions.}
  \label{fig:dvd}
\end{figure*}

\textbf{Qualitative Comparisons:} In Fig.~\ref{fig:dvd}, we show the restored images from different models. We note that state-of-the-art methods may fail to generate sharp images. In contrast, the proposed method restores many sharpened images. 

\textbf{Model Size and Inference Time:} As shown in Fig. \ref{fig:GoPro}, the single-scale alignment framework has an advanced performance on deblurring but brings more computational complexity. Therefore, we have proposed novel strategies to reduce the computational complexity. As shown in Table \ref{tab:main}, the number of parameters and runtime of our method is advantageous, which improves the performance of deblurring without a heavy computational burden.

\subsection{Model Analysis and Discussions}\label{section:exp}
In this section, we take the model trained on the GoPro ~\cite{nah2017deep} using three blurry frames for reconstruction to analyze the effects of multiple components in our framework.

\begin{figure}[t]

  \centering
  \begin{subfigure}{0.4\linewidth}
      \includegraphics[width=\textwidth]{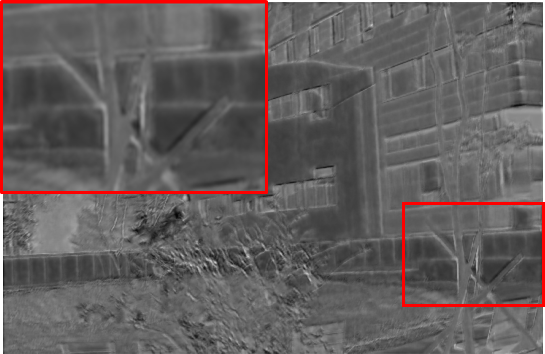}
    \caption{w/o SDD}

  \end{subfigure}
  \hspace{4mm}
  \begin{subfigure}{0.4\linewidth}
    \includegraphics[width=\textwidth]{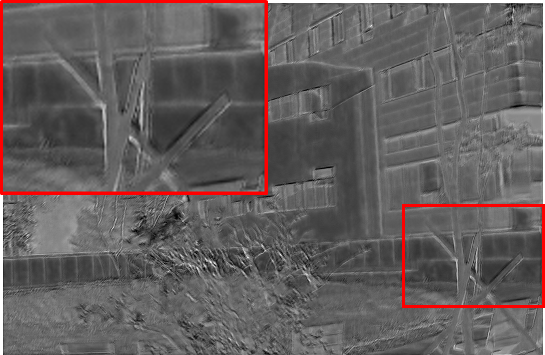}
  \caption{w/ SDD}
\end{subfigure}
\vspace{-3mm}
  \caption{Comparisons of feature map downsampled with SDD and without SDD.}
  \label{fig:SDD}
\end{figure}

\textbf{Effects of Structure-to-Detail Downsampling.}
In order to better enhance the edge of the features to facilitate the alignment of the features of the target and the neighboring frames, we design the Structure-to-Detail Downsampling. Compared with directly using a convolutional layer with a stride of 2 for downsampling, the performance gain of the SDD is shown in Table \ref{tab:SSD}. As shown in Fig.~\ref{fig:SDD}, using SDD can result in the objects in the downsampled feature map with sharper boundaries.
\begin{table}[t]
  \centering
  \caption{Effects of Structure-to-Detail Downsampling .}
    \begin{tabular}{ccc}
    \toprule
    Methods & w/o SDD & w/ SDD \\
    \midrule
    PSNR (dB) & 32.64 & 32.74 \\
    \bottomrule
    \end{tabular}%
  \label{tab:SSD}%
\vspace{-3mm}
\end{table}%
\begin{figure*}
  \centering
  \begin{subfigure}{0.185\textwidth}
  \begin{subfigure}{1\textwidth}
      \includegraphics[width=1\textwidth]{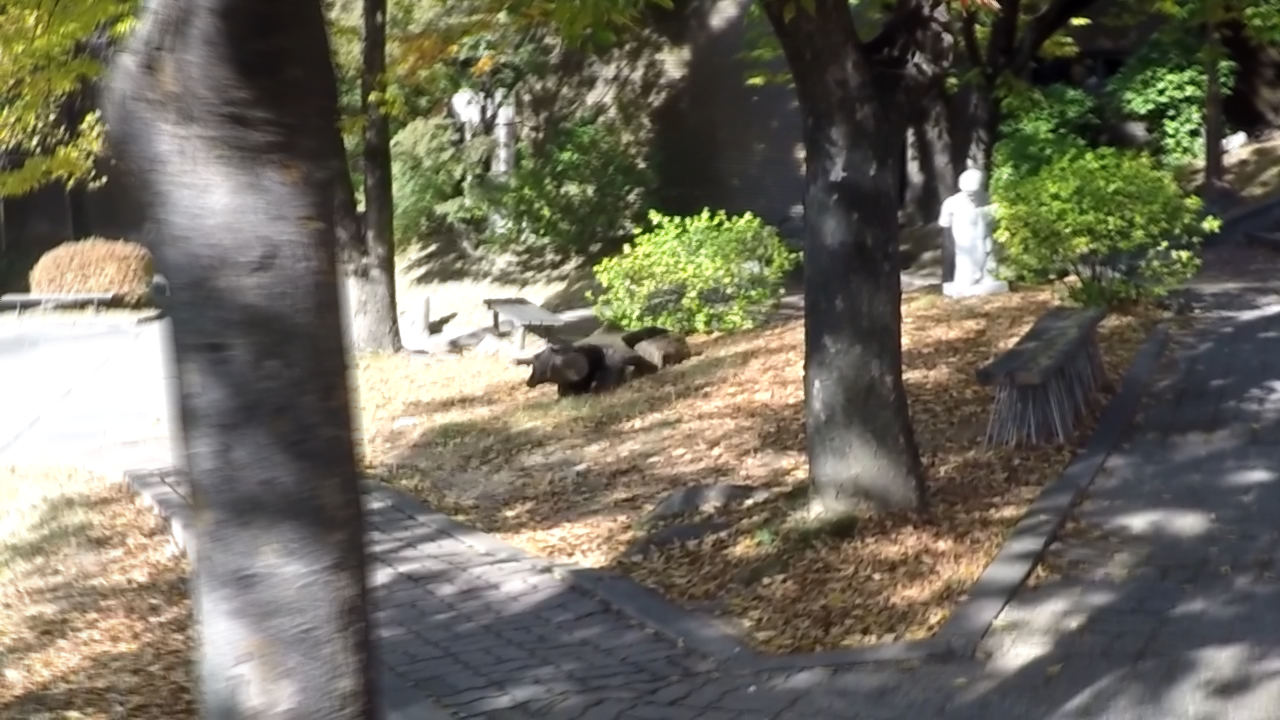}
    \subcaption*{Frame 47}
  \end{subfigure} \\
  \begin{subfigure}{1\textwidth}
    \includegraphics[width=1\textwidth]{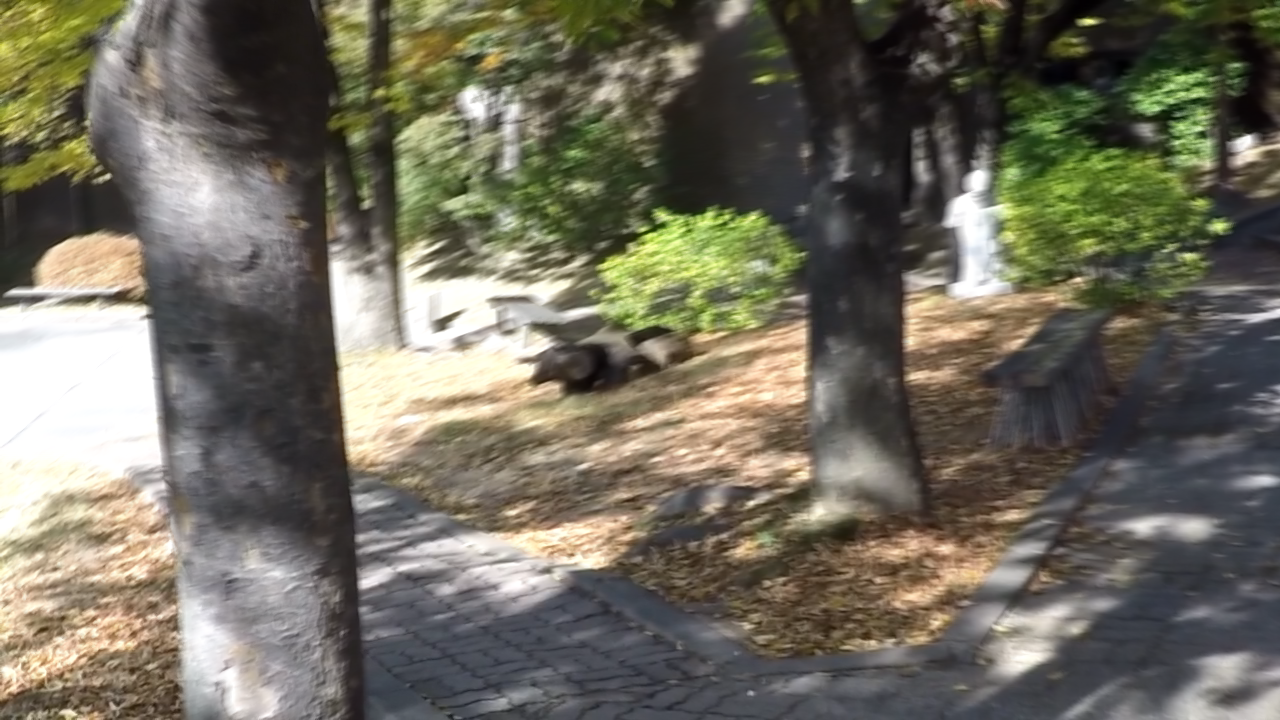}
    \subcaption*{Frame 48}
  \end{subfigure}
\end{subfigure}
\begin{subfigure}{0.185\textwidth}
    \includegraphics[width=1\textwidth]{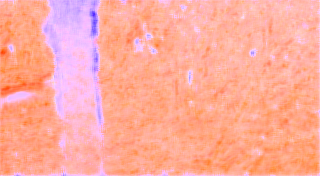} \\
    \includegraphics[width=1\textwidth]{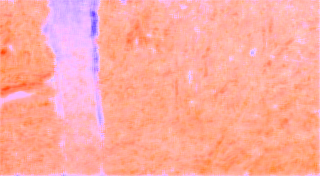} \\
    \includegraphics[width=1\textwidth]{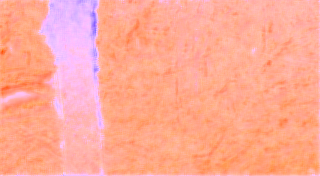} 
    \subcaption*{L3 offset}
\end{subfigure}
\begin{subfigure}{0.185\textwidth}
  \includegraphics[width=1\textwidth]{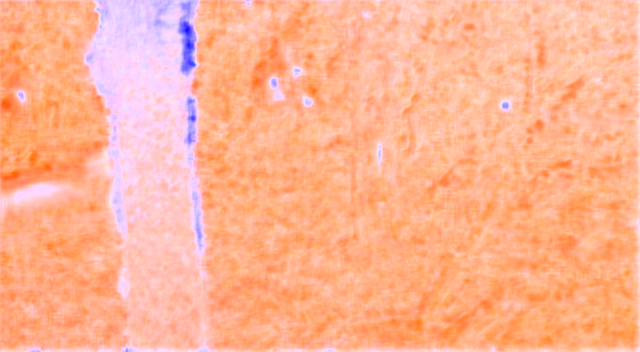} \\
  \includegraphics[width=1\textwidth]{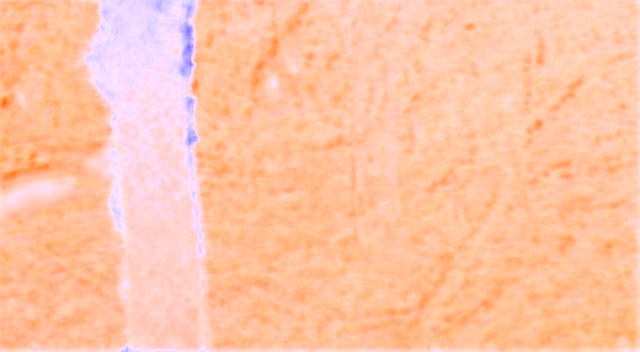} \\
  \includegraphics[width=1\textwidth]{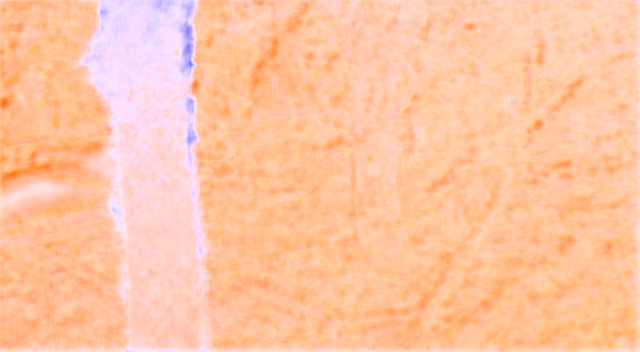}  
  \subcaption*{L2 offset}
\end{subfigure}
\begin{subfigure}{0.185\textwidth}
  \includegraphics[width=1\textwidth]{figures/offset/1/43_2.png} \\
  \includegraphics[width=1\textwidth]{figures/offset/12/43_1.png} \\
  \includegraphics[width=1\textwidth]{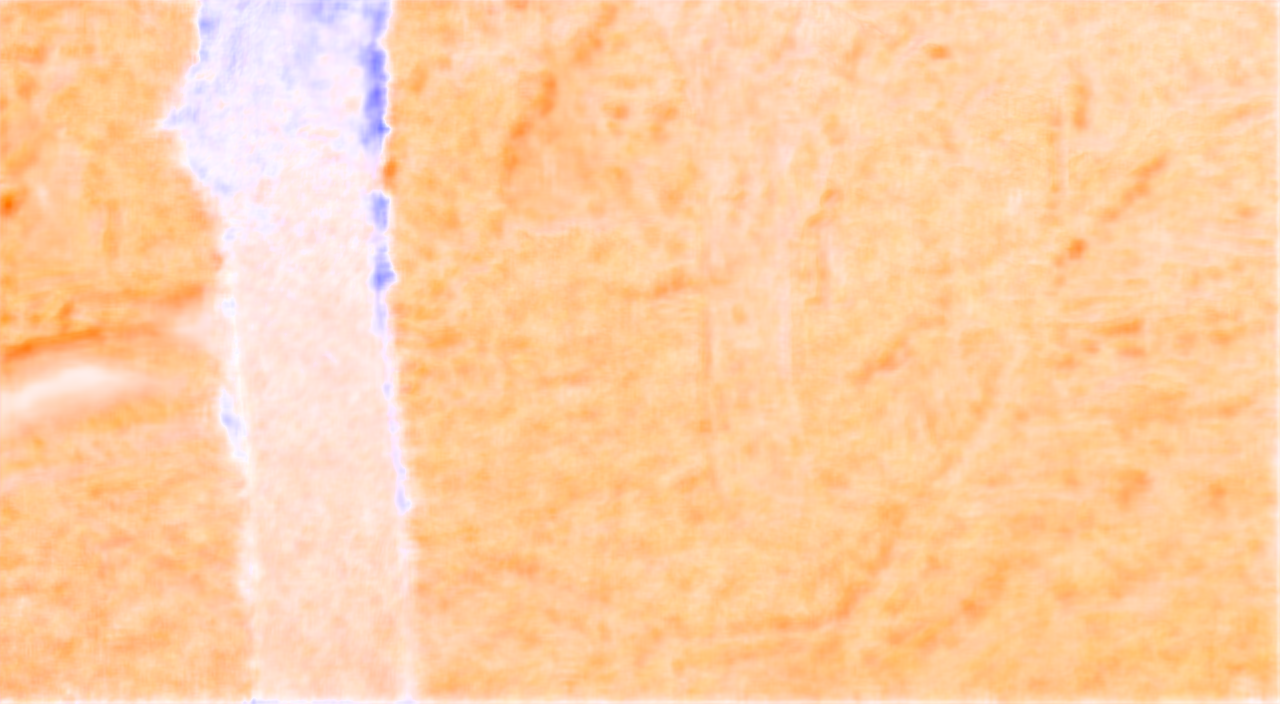}  
  \subcaption*{L1 offset}
\end{subfigure}
\begin{subfigure}{0.185\textwidth}
  \includegraphics[width=1\textwidth]{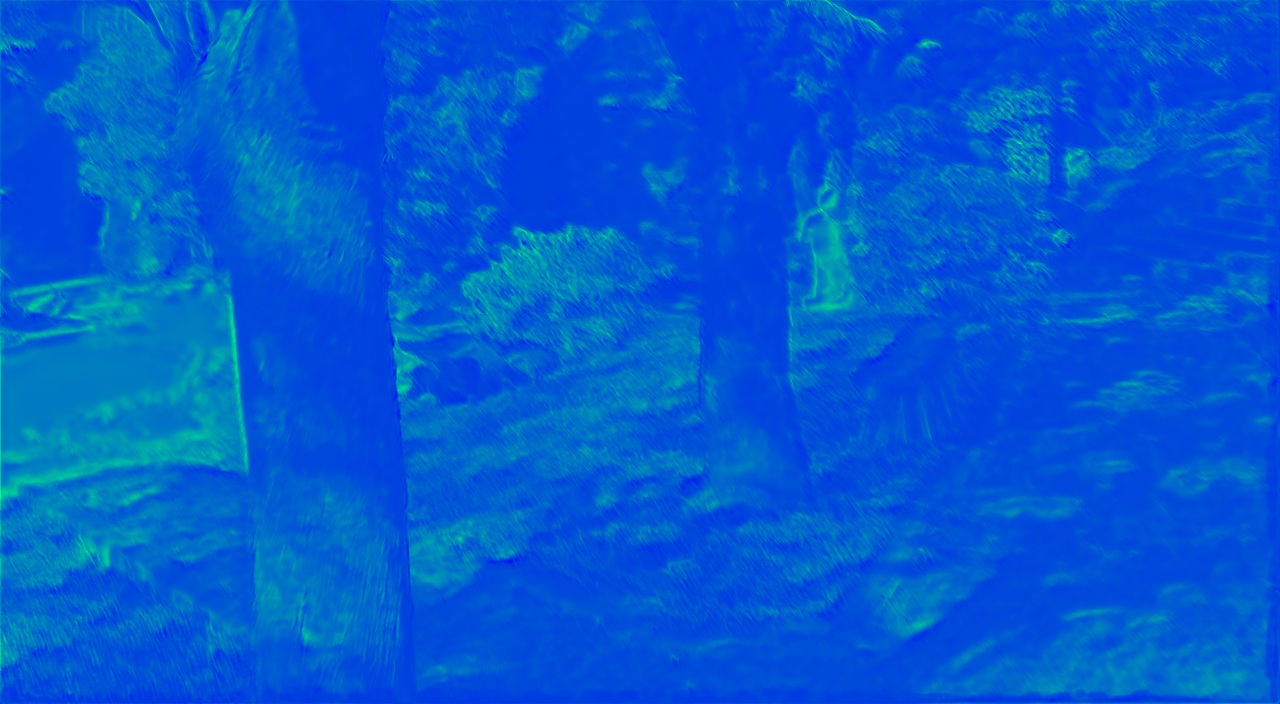} \\
  \includegraphics[width=1\textwidth]{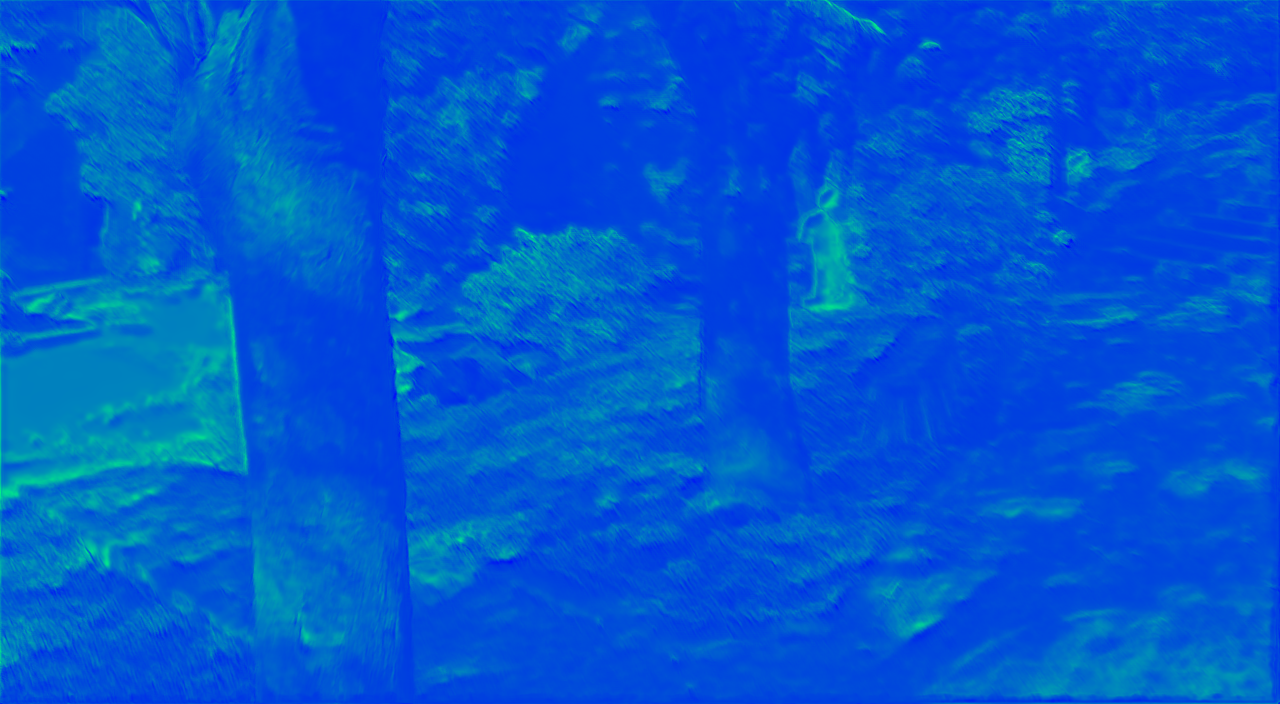} \\
  \includegraphics[width=1\textwidth]{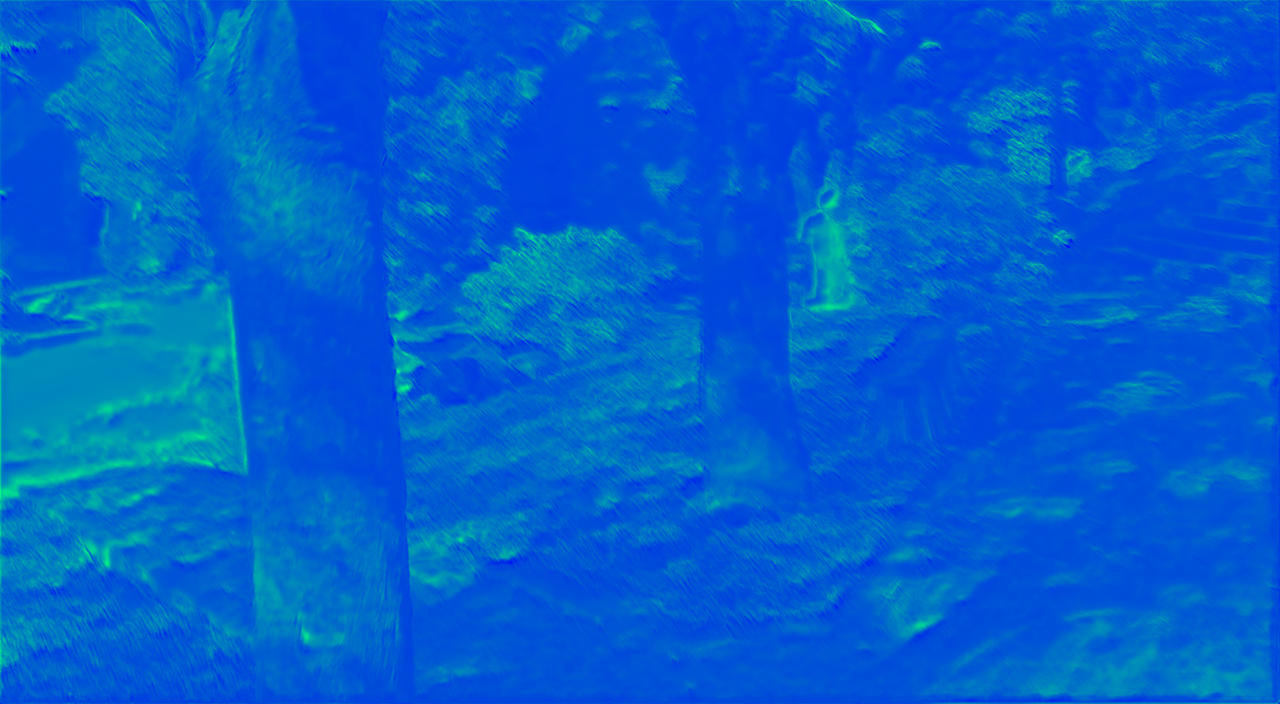}  
  \subcaption*{Aligned feature}
\end{subfigure}
\hspace{1mm}
\begin{subfigure}{0.02\textwidth}
  \rotatebox{90}{Type \uppercase\expandafter{\romannumeral4}~~~~~~~~Type \uppercase\expandafter{\romannumeral3}~~~~~~~~Type \uppercase\expandafter{\romannumeral1}} \\
\end{subfigure}
\caption{Effects of multi-scale supervision. Multi-scale supervision makes alignment modules to reduce errors and correct errors.}
\label{fig:offset}
\vspace{-4mm}
\end{figure*}

\textbf{Effects of Pyramid Reconstruction.}
In order to handle spatially invariant blur, we propose to construct a pyramid network to explore multi-scale information. To analyze the effects of the pyramid network, we experiment with different choices of pyramid layers with L = 2, 3, 4. As shown in Table \ref{tab:level}, learning features using a pyramid network improves the deblurring results. As the number of pyramid layers increases, we observe improvements in the restoration quality on account of
the enlarged receptive fields. Finally, taking both the performance and complexity into consideration, we choose the pyramid with three levels.
\begin{table}[t]
  \centering
  \caption{Effects of using the different numbers of pyramid level.}
    \begin{tabular}{cccc}
    \toprule
    Level & 2     & 3     & 4 \\
    \midrule
    PSNR (dB)  & 30.92 & 32.74 & 33.01 \\
    Parameters ($\times 10^{6}$)  & 2.12 & 7.94 & 32.07 \\
    Runtime (s)  & 0.02 & 0.20 & 0.46 \\
    \bottomrule
    \end{tabular}%
  \label{tab:level}%
  \vspace{-2mm}
\end{table}%

\begin{table}[t]
  \centering
  \caption{Effects of multi-scale supervision. The check mark indicates the supervision of reconstruction in the corresponding number of layers. For example, the presence of a ``$\surd$'' under column ``1'' means that the loss is imposed on reconstruction of images with the highest resolution.}
    \begin{tabular}{ccccc}
    \toprule
    Type  & 1     & 2     & 3     & PSNR (dB) \\
    \midrule
    \uppercase\expandafter{\romannumeral1}     & $\surd$  &       &       & 32.21 \\
    \uppercase\expandafter{\romannumeral2}     & $\surd$  &       & $\surd$  & 32.57 \\
    \uppercase\expandafter{\romannumeral3}     & $\surd$  & $\surd$  &       & 32.62 \\
    \uppercase\expandafter{\romannumeral4}     & $\surd$  & $\surd$  & $\surd$  & 32.74 \\
    \bottomrule
    \end{tabular}%
  \label{tab:loss}%
  \vspace{-3mm}
\end{table}

\textbf{Effects of Multiple Scale Supervision.} As we have mentioned in Section \ref{subsection:loss},  supervising multi-scale alignment can result in better reconstruction results. To empirically demonstrate the effectiveness of this strategy, we remove the low-scale supervision in the proposed framework with other settings unaltered for ablations. Specifically, we conducted four types of ablation training strategies for comparisons. Type \uppercase\expandafter{\romannumeral1} adopts the supervision only at level 1 of the pyramid. Type \uppercase\expandafter{\romannumeral2} adopts the supervision at both levels 1 and 3. Type \uppercase\expandafter{\romannumeral3} adopts the supervision at both levels 1 and 2.  Type \uppercase\expandafter{\romannumeral4} adopts the supervision at all levels.  As shown in Table \ref{tab:loss},  supervising the network at all scales improves the performance of video deblurring. To investigate why the performance is improved, we visualize the offsets at different scales. As shown in Fig. \ref{fig:offset}, compared to Type \uppercase\expandafter{\romannumeral4}, the model trained by Type \uppercase\expandafter{\romannumeral1} has errors at the coarsest motion estimation due to the lack of adequate supervision signals. Furthermore,  errors will accumulate into higher levels, which causes serious misalignment at the highest scale. In contrast, our studies show that sufficient supervision can reduce errors in motion estimation at the coarsest scale and bring the corresponding performance gain for deblurring. For both Type {\uppercase\expandafter{\romannumeral1}} and {\uppercase\expandafter{\romannumeral3}}, we notice that alignment errors exist in the motion estimation at level 3. However, with supervision introduced at level 2, the model trained according to Type \uppercase\expandafter{\romannumeral3} corrects some errors in the previously estimated motion. Compared to Type {\uppercase\expandafter{\romannumeral1}}, Type {\uppercase\expandafter{\romannumeral3}} prevents the errors from accumulation to higher scales. Meanwhile, only removing the third layer of supervision and keeping the second layer of supervision, alignment error still exists in the coarsest scale. However, due to the supervision from the second layer, some errors have been corrected. The intermediate supervision reduces the error accumulation and leads to less errors for the final offset. Therefore, multi-scale supervision reduces the errors at coarse scales and corrects the existing errors, which results in a performance gain for deblurring.

\begin{figure}[t]\footnotesize
  \centering
  \begin{tabular}{cccc}
  \includegraphics[width=0.32\linewidth]{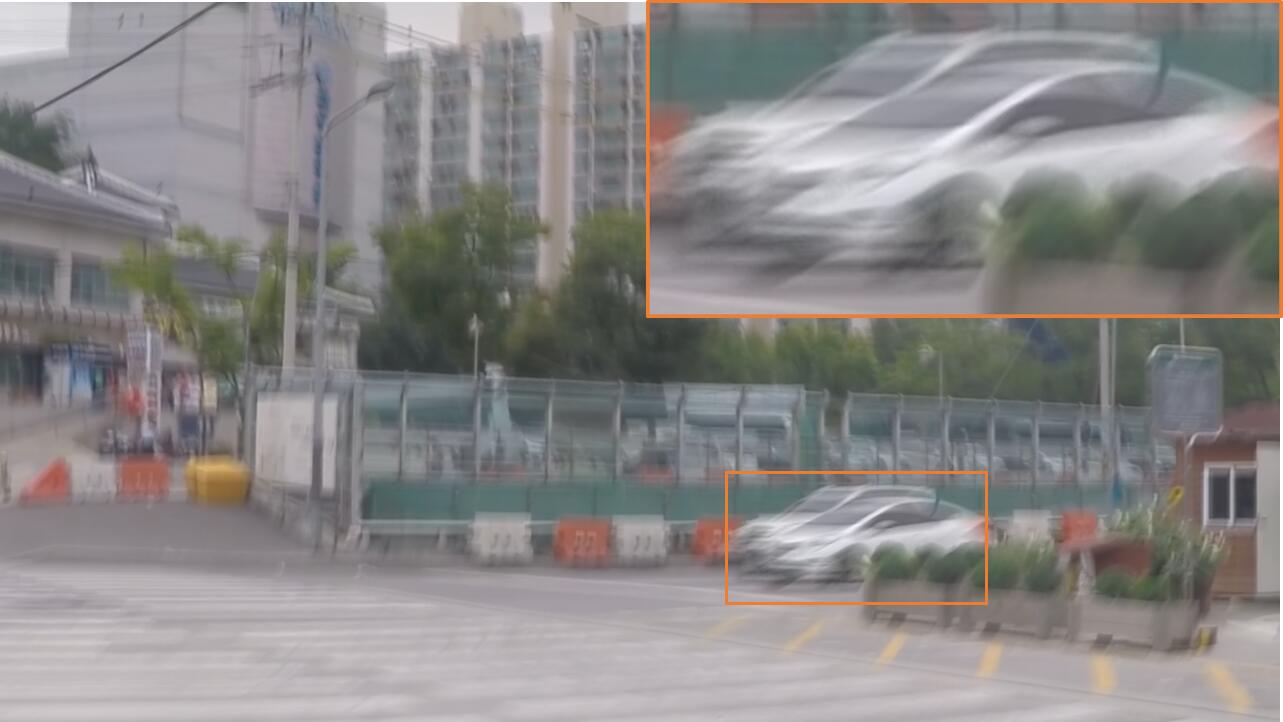} &\hspace{-4.5mm}
  \includegraphics[width=0.32\linewidth]{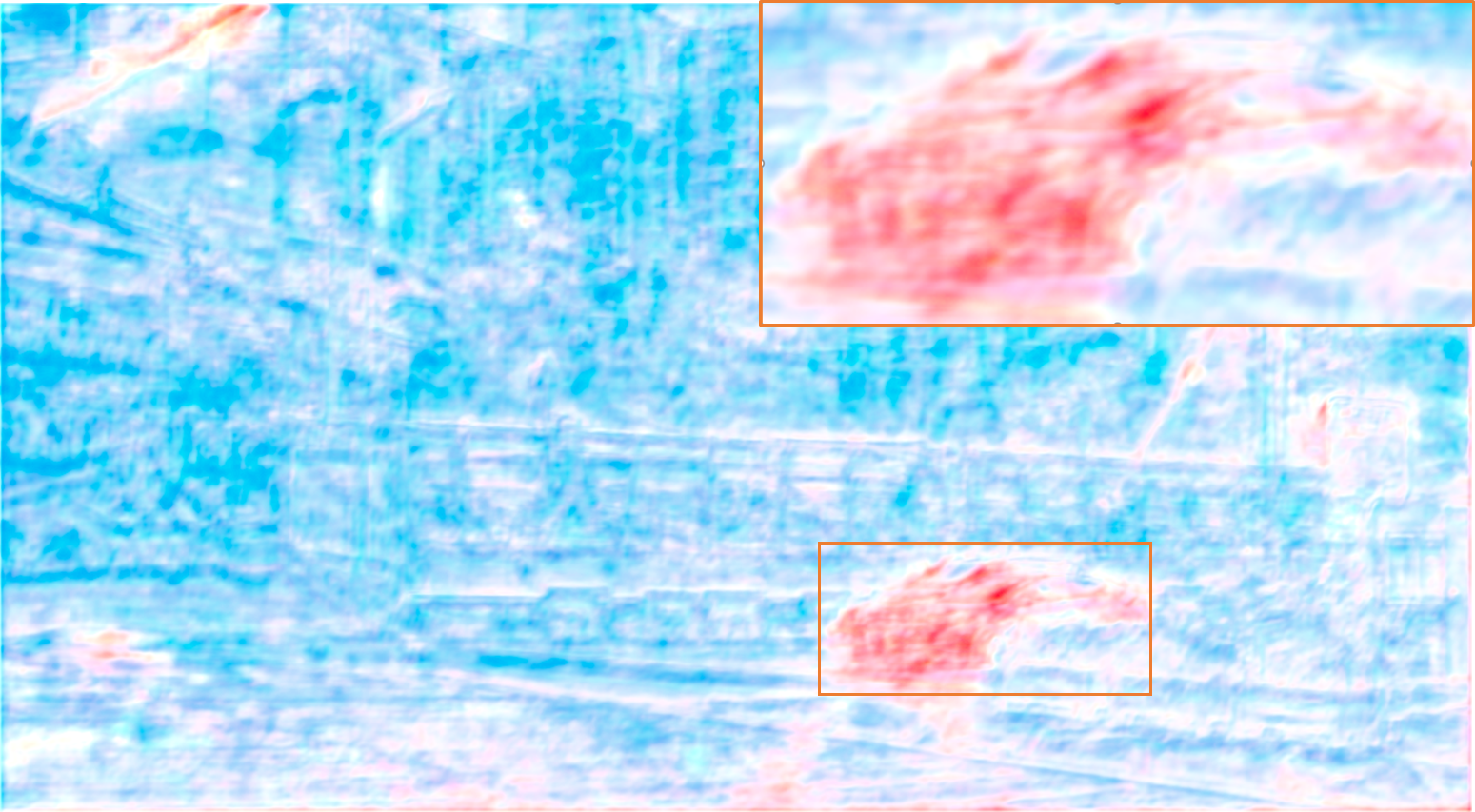}  &\hspace{-4.5mm}
  \includegraphics[width=0.32\linewidth]{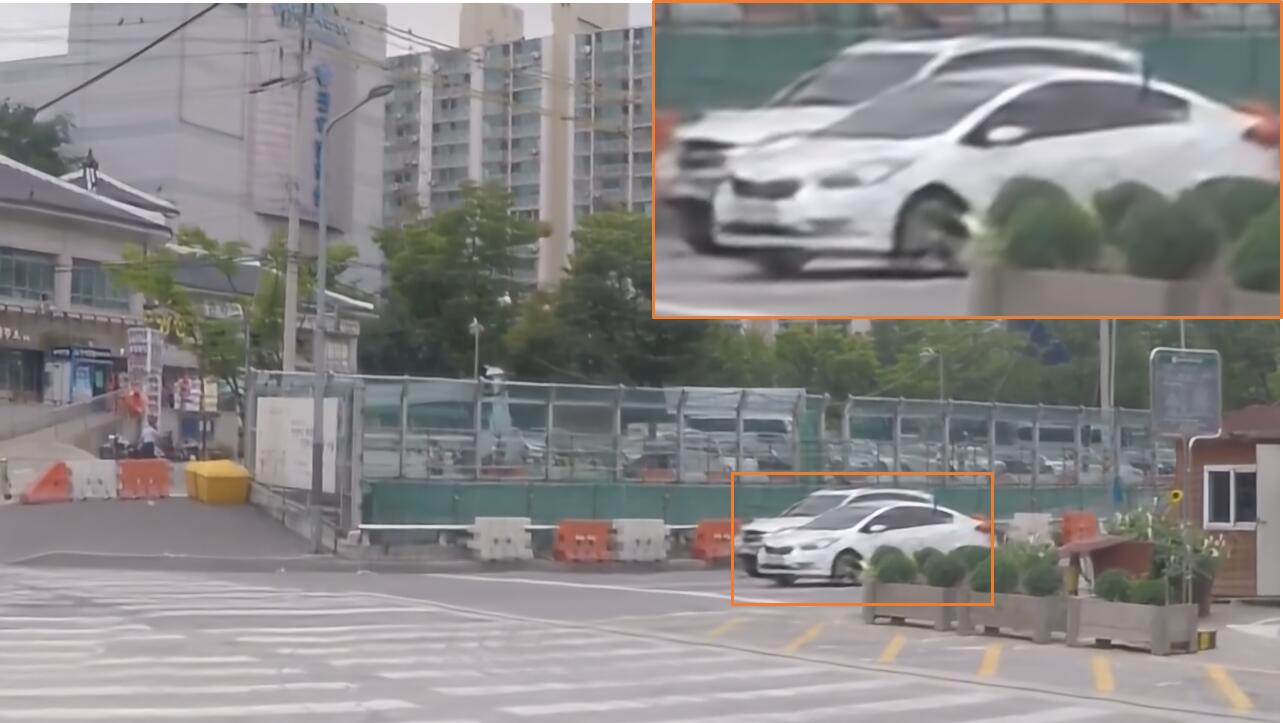} &\hspace{-4.5mm} \\
   (a) blurry frame &\hspace{-4mm}  (b) the offset of PCD &\hspace{-4mm}  (c) w/ PCD\\
  \includegraphics[width=0.32\linewidth]{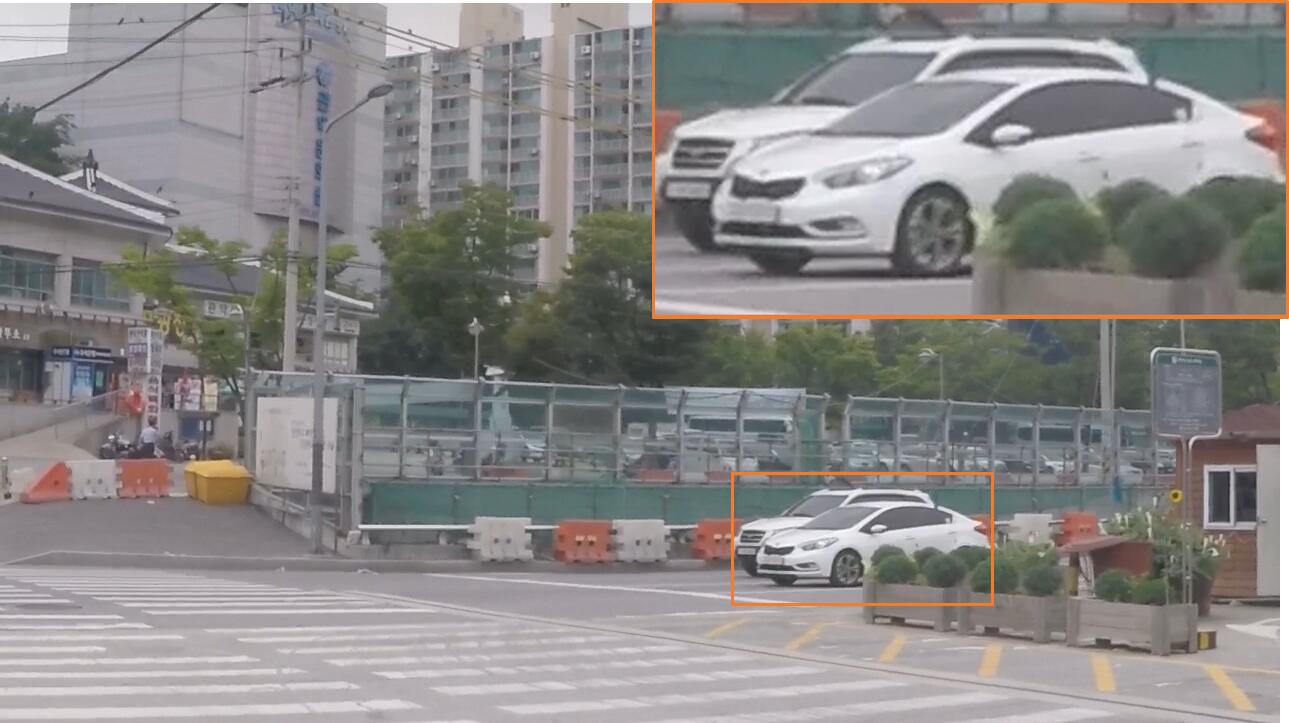}  &\hspace{-4.5mm}
  \includegraphics[width=0.32\linewidth]{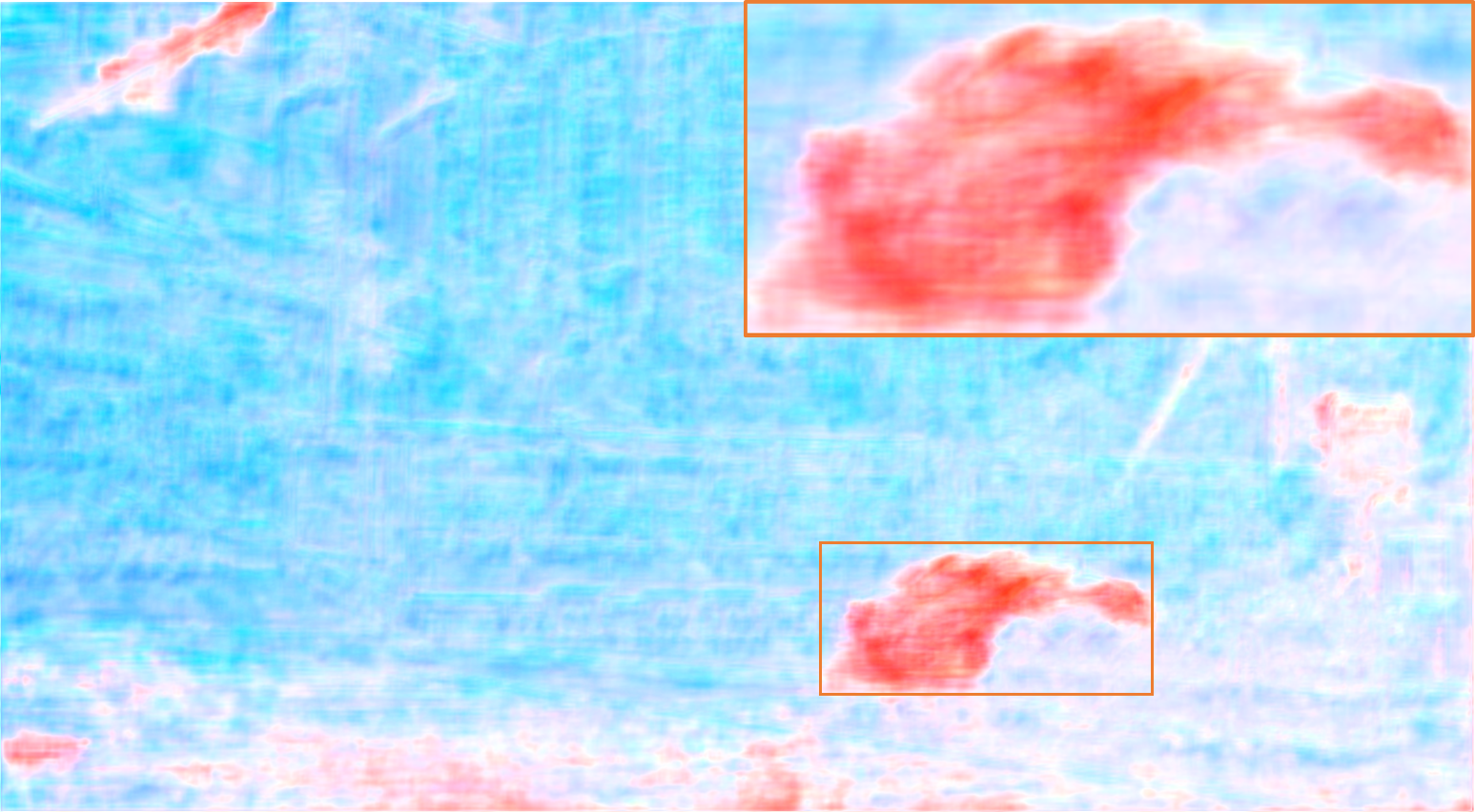}  &\hspace{-4.5mm}
  \includegraphics[width=0.32\linewidth]{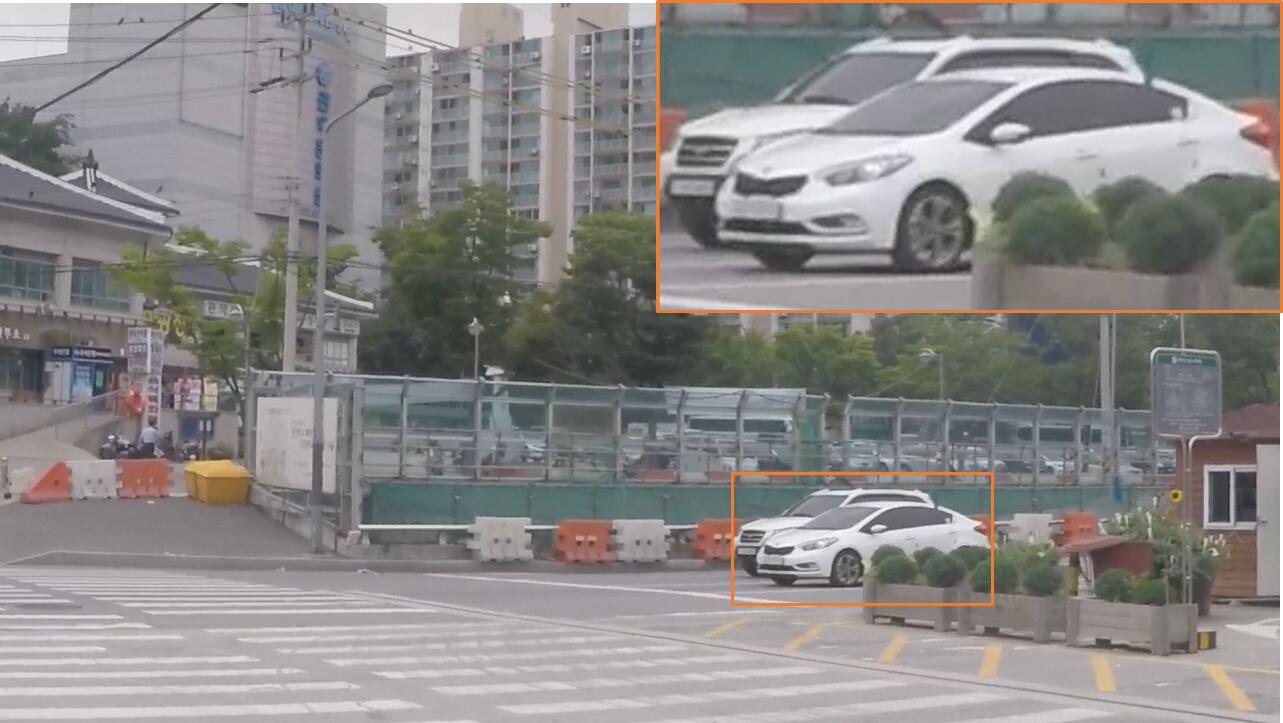}  &\hspace{-4.5mm} \\
 (d) GT &\hspace{-4mm}  (e) the offset of CGDA &\hspace{-4mm}  (f) w/ CGDA \\

  \end{tabular}

  \caption{Effects of alignment on video deblurring. Compared to PCD, the offset obtained by CGDA is smoother, so there are less artifacts in the recovered image.}
  \label{fig:flow}
  \vspace{-2mm}
  \end{figure}

\textbf{Effects of Alignment Methods.}
We conduct comparative experiments under three settings: (1) No alignment: No alignment. (2) 3-DCN: Each scale uses a layer of DCN for alignment and does not use coarse-grained motion estimation to guide fine-grained motion estimation. (3) PCD: Performing the alignment with Pyramid, Cascading, and Deformable Convolution~\cite{wang2019edvr}. (4) CGDA: Adopting our proposed CGDA. Compared to misalignment, using the simplest 3-DCN can effectively improve PSNR by 0.33 dB. However, due to the large and complex motion between frames, it is challenging to perform alignment at a high scale with one-layer DCN directly. Moreover, the performance of simply stacking three DCN layers is still unsatisfactory. Therefore, compared with 3-DCN, PCD that tries to generate offset by considering the fused feature achieves a gain of 0.30 dB on PSNR. Compared with implicitly fusing the multi-scale in PCD, we explicitly consider the offset of low-scale estimation as course offset to pre-align the feature. Consequently, estimating the motion at a large scale only needs to generate residual offset. This strategy improve the stability of training with reduced parameters. The network trained with CGDA achieves a gain of 0.18 dB on PSNR compared to the PCD. Fig.~\ref{fig:flow} demonstrates that the estimated offset by PFAN is smoother, which enhances the quality of the restored frame.

\begin{table}[t]
  \centering
  \caption{Effects of using different alignment methods for PFAN.}
    \begin{tabular}{ccccc}
    \toprule
    Methods & no align & 3 DCN & PCD & CGDA \\
    \midrule
    PSNR (dB)  & 31.93 & 32.26 & 32.56 & 32.74 \\
    Parameters ($\times 10^{6}$) & - & 1.02 & 1.18 & 1.09 \\
    \bottomrule
    \end{tabular}%
  \label{tab:Align}%
  \vspace{-2mm}
\end{table}%

\section{Conclusions}
We have developed a simple yet effective framework that uses the classic coarse-to-fine principle for video deblurring. By enhancing the supervision of multi-scale motion estimation, our proposed framework effectively reduces the misalignment of features. The low-scale motion information is used to guide large-scale motion information to realize multi-layer alignment with a simple structure. The experimental results demonstrate that our pyramid framework outperforms the traditional single-scale alignment framework regarding the computational complexity and performance trade-off.
\newpage
{\small
\bibliographystyle{ieee_fullname}
\bibliography{egbib}
}

\end{document}